\documentclass[10pt,twocolumn,letterpaper]{article}

\usepackage[pagenumbers]{cvpr} 

\usepackage{graphicx}
\usepackage{amsmath}
\usepackage{amssymb}
\usepackage{pifont}
\usepackage{booktabs}
\usepackage{multirow}
\usepackage[accsupp]{axessibility}
\newcommand{\cmark}{\text{\ding{51}}}
\newcommand{\xmark}{\text{\ding{55}}}
\usepackage[table,x11names]{xcolor}
\definecolor{maroon}{cmyk}{0,0.87,0.68,0.32}

\usepackage[pagebackref,breaklinks,colorlinks]{hyperref}

\usepackage[capitalize]{cleveref}
\crefname{section}{Sec.}{Secs.}
\Crefname{section}{Section}{Sections}
\Crefname{table}{Table}{Tables}
\crefname{table}{Tab.}{Tabs.}

\def\cvprPaperID{1207} 
\def\confName{CVPR}
\def\confYear{2023}

\begin{document}

\title{Visual Language Pretrained Multiple Instance Zero-Shot Transfer for Histopathology Images}


\author{Ming Y. Lu$^{\dagger, 1,2,3}$, Bowen Chen$^{\dagger, 2,3}$, Andrew Zhang$^{1,2,3}$, Drew F.K. Williamson$^{2,3}$, \\ Richard J. Chen$^{2,3}$, Tong Ding$^{2,3}$, Long Phi Le$^{2,3}$, Yung-Sung Chuang$^{1}$, Faisal Mahmood$^{2,3}$\\
$^{1}$Massachusetts Institute of Technology \enspace 
$^{2}$Harvard University \enspace $^{3}$Mass General Brigham \enspace\\
{\tt\small {mingylu@mit.edu, bchen18@bwh.harvard.edu, faisalmahmood@bwh.harvard.edu}}
}

\maketitle
\def\thefootnote{$\dagger$}\footnotetext{These authors contributed equally to this work.}\def\thefootnote{\arabic{footnote}}

\begin{abstract}
Contrastive visual language pretraining has emerged as a powerful method for either training new language-aware image encoders or augmenting existing pretrained models with zero-shot visual recognition capabilities. However, existing works typically train on large datasets of image-text pairs and have been designed to perform downstream tasks involving only small to medium sized-images, neither of which are applicable to the emerging field of computational pathology where there are limited publicly available paired image-text datasets and each image can span up to 100,000 $\times$ 100,000 pixels. In this paper we present MI-Zero,  a simple and intuitive framework for unleashing the zero-shot transfer capabilities of contrastively aligned image and text models on gigapixel histopathology whole slide images, enabling multiple downstream diagnostic tasks to be carried out by pretrained encoders without requiring any additional labels. MI-Zero reformulates zero-shot transfer under the framework of multiple instance learning to overcome the computational challenge of inference on extremely large images. We used over 550k pathology reports and other available in-domain text corpora to pretrain our text encoder. By effectively leveraging strong pretrained encoders, our best model pretrained on over 33k histopathology image-caption pairs achieves an average median zero-shot accuracy of 70.2\% across three different real-world cancer subtyping tasks. Our code is available at: \href{https://github.com/mahmoodlab/MI-Zero}{https://github.com/mahmoodlab/MI-Zero}.
\end{abstract}


\vspace{-3mm}
\section{Introduction}
\label{sec:intro}


Weakly-supervised deep learning for computational pathology (CPATH) has rapidly become a standard approach for modelling whole slide image (WSI) data~\cite{ilse2018attention, campanella2019clinical, topol2019high,  van2021deep, lu2021ai}. To obtain ``clinical grade" machine learning performance on par with human experts for a given clinical task, many approaches adopt the following model development life cycle: 1) curate a large patient cohort ($N > 1000$ samples) with diagnostic whole-slide images and clinical labels, 2) unravel and tokenize the WSI into a sequence of patch features, 3) use labels to train a slide classifier that learns to aggregate the patch features for making a prediction, and 4) transfer the slide classifier for downstream clinical deployment~\cite{campanella2019clinical, zhao2020predicting, li2021dual}. 


Successful examples of task-specific model development (\textit{e.g.} training models from scratch for each task) include prostate cancer grading and lymph node metastasis detection~\cite{bejnordi2017diagnostic, campanella2019clinical, bulten2020automated, bulten2022artificial, tolkach2020high, nagpal2019development}. However, this paradigm is intractable if one wishes to scale across the hundreds of tumor types across the dozens of different organ sites in the WHO classification system\footnote{\href{https://tumourclassification.iarc.who.int/}{tumourclassification.iarc.who.int/}}, with most tumor types under-represented in public datasets or having inadequate samples for model development~\cite{kundra2021oncotree, zhu2017wsisa}. To partially address these limitations, self-supervised learning has been explored for learning the patch representations within the WSI with the idea that certain local features, such as tumor cells, lymphocytes, and stroma, may be conserved and transferred across tissue types~\cite{ciga2022self, wang2022transformer, koohbanani2021self, li2021sslp, krishnan2022self, srinidhi2021improving, chen2022scaling}. Though morphological features at the patch-level are captured in a task-agnostic fashion, developing the slide classifier still requires supervision, which may not be possible for disease types with small sample sizes. To scale slide classification across the vast number of clinical tasks and possible findings in CPATH, an important shift needs to be made from task-specific to task-agnostic model development.



Recent works \cite{clip,jia2021scaling} have demonstrated that large-scale pretraining using massive, web-sourced datasets of noisy image-text pairs can not only learn well-aligned representation spaces between image and language, but also transfer the aligned latent space to perform downstream tasks such as image classification. Specifically for CLIP \cite{clip}, after pretraining a vision encoder in a task-agnostic fashion, the vision encoder can be ``prompted" with text from the label space (referred to as ``zero-shot transfer", as no labeled examples are used in the transfer protocol). Despite the volume of zero-shot transfer applications developed for natural images~\cite{radford2019language, jia2021scaling, wu2021data, furst2021cloob, mu2022slip, li2021supervision, yang2022vision} and certain medical imaging modalities (\textit{e.g.} radiology~\cite{zhang2022contrastive, tiu2022expert, sellergren2022simplified,  huang2021gloria, wang2022medclip}), zero-shot transfer for pathology has not yet been studied\footnote{Concurrent to our work, BiomedCLIP\cite{zhang2023large} was developed using figure-caption pairs mined from PubMed articles. It was benchmarked on both patch-level histopathology datasets and radiology datasets, but did not study zero-shot transfer for gigapixel WSIs.}. We believe this is due to 1) the lack of large-scale, publicly available datasets of paired images and captions in the highly specialized field of pathology, and 2) fundamental computational challenges associated with WSIs, as images can span up to $100,000 \times 100,000$ pixels and do not routinely come with textual descriptions, bounding box annotations or even region of interest labels.

In this work, we overcome the above data and computational challenges and develop the first zero-shot transfer framework for the classification of histopathology whole slide images. On the data end, we curated the largest known dataset of web-sourced image-caption pairs specifically for pathology. 
We propose ``MI-Zero", a simple and intuitive multiple instance learning-based \cite{amores2013multiple, ilse2018attention} method for utilizing the zero-shot transfer capability of pretrained visual-language encoders for gigapixel-sized WSIs that are routinely examined during clinical practice. We validate our approach on 3 different real-world cancer subtyping tasks, and perform multiple ablation experiments that explore image pretraining, text pretraining, pooling strategies, and sample size choices for enabling zero-shot transfer in MI-Zero.





\begin{figure*}
  \centering
   \includegraphics[width=1\linewidth]{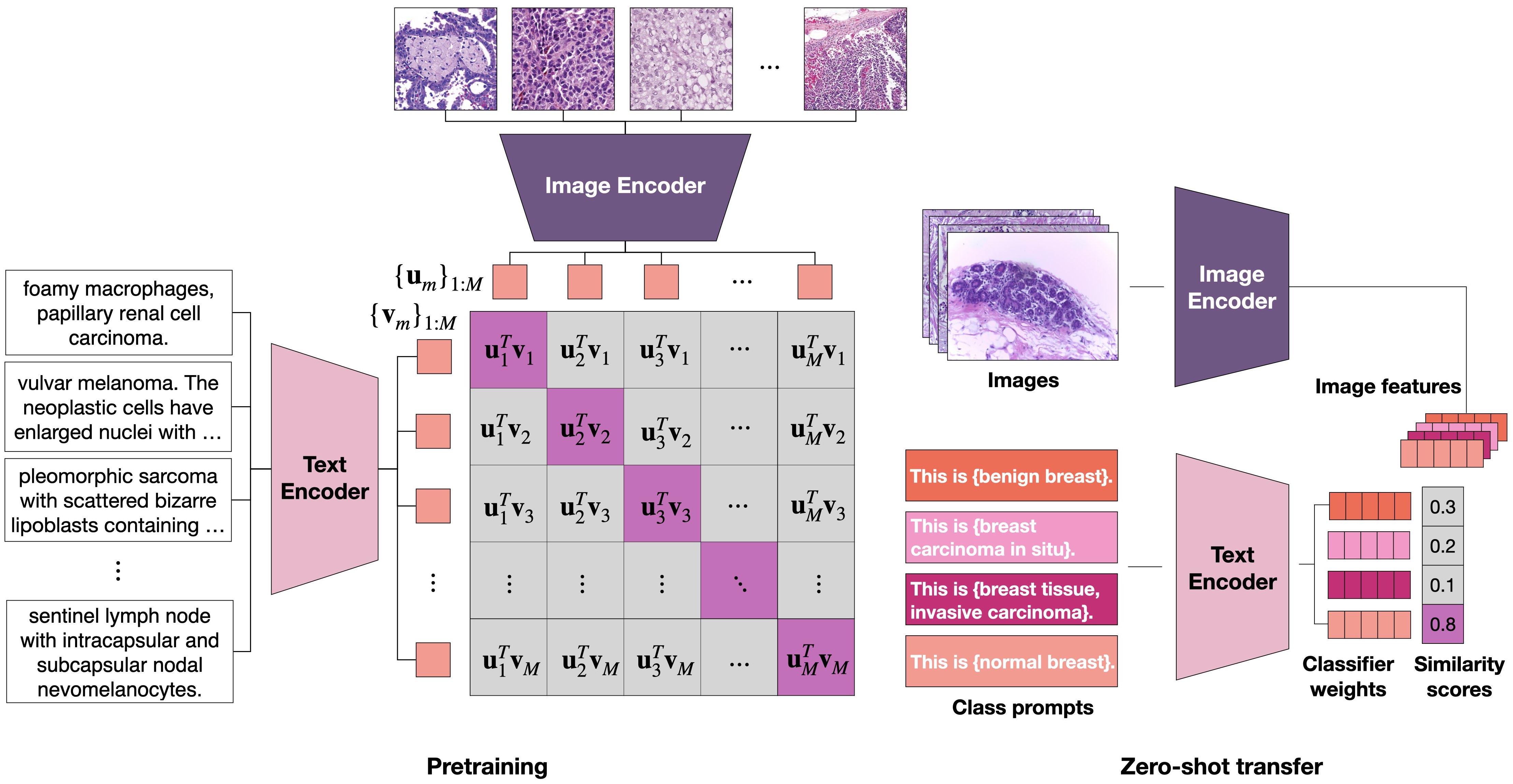}
   \caption{Illustration of contrastive \textbf{visual language pretraining} (left). Visual and language embeddings are aligned using a cross-modal contrastive loss. For our experiments, we curated the largest image-caption dataset in pathology consisting of 33,480 pairs. \textbf{Zero-shot transfer} for image classification (right), which we generalize for slide-level classification using MI-Zero (See \Cref{fig:wsi}).}
   \label{fig:patch}
\end{figure*}

\vspace{-0.1cm}
\section{Related Work}
\label{sec:related}

\noindent \textbf{Contrastive visual representation learning}. Contrastive learning has emerged as a powerful pretraining technique for learning task-agnostic representations of data. It works by constructing collections of similar samples (positive pairs) that would have embeddings maximally aligned in the model's latent space, as well as dissimilar samples (negative pairs), for which embeddings should be spread far apart\cite{zou2013contrastive, gutmann2010noise, oord2018representation, chen2020simple}. Examples from computer vision include different augmented views of the same unlabeled image\cite{chen2020simple, he2020momentum}, images with the same class label~\cite{khosla2020supervised, yang2022unified, fashi2022self}, and different sensory views of the same scene~\cite{cmc}. In computational pathology, recent works\cite{ciga2022self, wang2022transformer, koohbanani2021self, srinidhi2021improving, vu2023handcrafted, an2022masked} have leveraged contrastive learning and unlabeled images from histopathology datasets to pretrain domain-specific image encoders that achieve strong performance on downstream visual recognition tasks compared to transfer learning.

\vspace{3mm}
\noindent \textbf{Visual language pretraining}. Contrastive learning has also been shown to be a highly effective and scalable strategy to pretrain dual-encoder image-text models that can excel at a range of downstream visual recognition tasks. In medical imaging, ConVIRT~\cite{zhang2022contrastive} considered paired chest X-ray images and reports for learning aligned visual language representation. Later, representative works such as CLIP~\cite{radford2021learning} and ALIGN~\cite{jia2021scaling} showed that by scaling to large, diverse web-source datasets of paired images and captions, we can train models capable of exhibiting fairly robust zero-shot transfer capabilities through the use of prompts that exploit the cross-modal alignment between image and text learned by the model during pretraining. Recent methods have explored ways to improve the sample efficiency and zero-shot performance of CLIP-like models~\cite{wu2021data, huang2021gloria, li2021supervision, furst2021cloob, mu2022slip, yang2022unified, lit}. Other works have also explored incorporating a generative loss and masked language modeling loss into the pretraining objective either in addition to or in lieu of contrastive-based objectives~\cite{wang2021simvlm, desai2021virtex, piergiovanni2022answer, Yu2022CoCaCC, lu2019vilbert, wang2022image, sariyildiz2020learning}.  Notably, VirTex\cite{desai2021virtex} has been used to learn visual representations for histopathology images using a generative captioning loss and the ARCH dataset~\cite{gamper2021multiple} containing 7,562\footnote{The number differs from \cite{gamper2021multiple} due to removing a few empty (`` "), invalid (\textit{e.g.} ``\texttt{(continued)}") or unpaired captions.} histopathology image-caption pairs from pathology textbooks and PubMed research articles.


\vspace{2mm}
\noindent \textbf{Multiple instance learning.} Multiple instance learning (MIL)\cite{amores2013multiple} refers to a family of methods that considers learning from weakly-annotated data where each input is a bag or collection of instances such that only an unknown subset of instances are relevant to or representative of the label. In CPATH, algorithms based around the framework of MIL have been proposed for various diagnostic tasks in a weakly-supervised setting, utilizing trainable aggregation operators to learn WSI-level embeddings independent of the size of the bag~\cite{ilse2018attention, campanella2019clinical}. ABMIL~\cite{ilse2018attention} proposed to use attention-guided weighted averaging as a generic operator to aggregate instance-level embeddings. CLAM\cite{lu2021data} took a first step towards data-efficient weakly-supervised learning for WSIs by embedding instances using a frozen ResNet encoder pretrained on ImageNet. Other variants and extensions of the vanilla MIL and ABMIL formulations have been studied\cite{javedadditive, zhang2022dtfd, yufeibayes, xiang2023exploring, hou2016patch, wangscl}, including extensions with self-supervised learning\cite{li2021dual, tellez2019neural, chen2022scaling, zhao2020predicting}, multi-scale feature aggregation\cite{li2021dual, zhang2021joint, tokunaga2019adaptive, chen2022scaling}, graphs\cite{zhao2020predicting, pati2022hierarchical, di2022generating, chen2021whole, lee2022derivation}, Transformer attention\cite{kalra2020learning, shao2021transmil, chen2022scaling}, learning to zoom\cite{thandiackal2022differentiable, kong2022efficient
, bergner2023iterative}, and multi-modal fusion\cite{chen2021multimodal, chen2022pan}.

\section{Methods}
\label{sec:prelims}
\subsection{Image caption dataset.}

\interfootnotelinepenalty=10000
For this work, we curated a histopathology dataset of image-caption pairs by scraping from publicly available educational resources and incorporating the existing ARCH dataset. We perform cleaning and filtering, yielding a highly diverse dataset of 33,480 image-caption pairs covering a diverse set of tissue sites and morphologies (See \textbf{Supplementary Materials} for additional details). 





\subsection{Unsupervised pretraining of unimodal encoders.}
\label{subsec:textpretraining}
While our paired dataset currently represents the largest of its kind in the domain of histopathology, it is still considerably smaller than what is feasible in the domain of radiology (\textit{e.g.} MIMIC-CXR \cite{johnson2019mimic}, 217k pairs) and what is used in representative works in general machine learning (\textit{e.g.} LiT~\cite{lit}, 4B pairs, ALIGN~\cite{jia2021scaling}, 1.8B pairs, CLIP~\cite{clip}, 400M pairs). Therefore, we initialize our encoders using pretrained weights before aligning their latent space using paired examples. For the text encoder, we collected a corpus specific to the domain of pathology, which notably includes the final diagnosis section of over 550k surgical pathology reports from Massachusetts General Hospital and over 400k histopathology-relevant PubMed abstracts. In-house diagnostic reports were cleaned and de-identified using regex. We pretrain a GPT-style autoregressive transformer (following the same architecture as GPT2-medium \cite{radford2019language}) as the text encoder and will refer to it as HistPathGPT henceforth. Specifically, given a sequence of $T$ word tokens $w_1, \ldots, w_T$, it is augmented to a sequence of length $T + 2$: $\mathbf{t} = (\text{[BOS]}, w_1, \ldots, w_T, \text{[EOS]})$. We maximize the log-likelihood of each token under an autoregressive generative model parameterized by $\phi$:
\begin{equation}
   \mathcal{L}_\mathrm{clm}(\phi)= - \sum_{t=1}^{T+1} \log p\left(w_t \mid w_{0: t-1}; \phi\right) 
\end{equation}
Additionally, we also explore initializing from publicly available text encoders that have been trained on biomedical and clinical corpora such as PubMed abstracts and MIMIC \cite{johnson2016mimic}. We consider two pretrained models that fall into this category: BioClinicalBert~\cite{alsentzer2019publicly} and PubMedBert~\cite{gu2021domain}. 
For the image encoder, we explore 2 strategies: 1) initializing from ImageNet pretrained weights, and 2) using a SOTA publicly available encoder trained using self-supervised representation learning on a total of 15.5M unlabeled histopathology image patches \cite{wang2022transformer}. 

\subsection{Aligning vision and language embeddings.}
\label{subsec:vl}
We align the latent space of our visual and language encoders using the cross-modal contrastive loss formulated as a temperature scaled $M$-way classification\cite{sohn2016improved}, where $M$ is the global batch-size of image-text pairs participating in the loss computation. Similar or analogous formulations of the contrastive loss are widely used for both self-supervised representation learning~\cite{cmc, chen2020simple} and visual-language pretraining~\cite{jia2021scaling, clip, zhang2022contrastive}. Given a batch of $M$ paired image and text samples $\{(\mathbf{x}_m, \mathbf{t}_m)\}_{m=1,\ldots,M}$, $\ell_2$-normalized visual and text embeddings are computed via the visual and text encoders $f(\cdot; \theta)$ and 
$g(\cdot; \phi)$ respectively as $\mathbf{u}_m = \frac{f(\mathbf{x}_m; \theta)}{\lVert f(\mathbf{x}_m; \theta) \rVert}$ and $\mathbf{v}_m = \frac{g(\mathbf{t}_m; \phi)}{\lVert g(\mathbf{t}_m, \phi) \rVert}$.
The two directions of contrastive learning ($i \xrightarrow{} t$) and $t \xrightarrow{} i$) are viewed as symmetric and used jointly (with equal weight) to optimize the model during training, where $\tau$ is a temperature parameter:
\begin{equation}
\mathcal{L}_{i 2 t}(\theta, \phi)=-\sum_{i=1}^{M} \log \frac{\exp \left(\tau \boldsymbol{u}_i^{T} \boldsymbol{v}_i\right)}{\sum_{j=1}^{M} \exp \left(\tau \boldsymbol{u}_i^{T}  \boldsymbol{v}_j\right)}
\end{equation}

\begin{equation}
\mathcal{L}_{t 2 i}(\theta, \phi)=-\sum_{j=1}^{M} \log \frac{\exp \left(\tau \boldsymbol{v}_j^{T}  \boldsymbol{u}_j\right)}{\sum_{i=1}^{M} \exp \left(\tau \boldsymbol{v}_j^{T}  \boldsymbol{u}_i\right)}
\end{equation}
For a batch of $M$ image-text pairs, we see the connection to the aforementioned $M$-way classification problem by interpreting for each image (text), the index of its paired text (image) as the ground truth ``target", and the remaining $M - 1$ indices, which correspond to other texts (images) in the mini-batch as ``negatives". Each direction of the contrastive loss is then equivalent to using the temperature scaled cosine similarity scores between embeddings as predicted logits, and minimizing the cross-entropy loss. 


\subsection{Zero-shot transfer for image classifcation.}
We briefly describe the prompt-based approach to zero-shot classification popularized by ~\cite{clip}. For each class of interest, a prompt has two components, the classname (\textit{e.g.} \texttt{"adenocarcinoma"}) and the template (\textit{e.g.} \texttt{"an image showing \{\}."}), which collectively form the sequence of word tokens (\textit{e.g.} \texttt{"an image showing adenocarcinoma."}) that are embedded by the trained text encoder to form the weights of a linear classifier. Formally, for an image $\mathbf{x}$, we compute its $\ell_2$-normalized image embedding $\mathbf{u}$ using the image encoder. Given prompts $\{\textbf{t}_m\}_{m=1,\ldots, C}$ where $C$ is the total number of classes, the text encoder produces prompt embeddings $\{\mathbf{w}_m\}_{m=1,\ldots, C}$ where $\mathbf{w}_m = \frac{g(\textbf{t}_m;\phi)}{\lVert g(\textbf{t}_m;\phi) \rVert}$. The classification decision of the model is:
\begin{equation}
\hat{y} = \operatorname*{argmax}_{m} {\mathbf{u}^T \mathbf{w}_m}
\end{equation}

\begin{figure*}
  \centering
   \includegraphics[width=.95\linewidth]{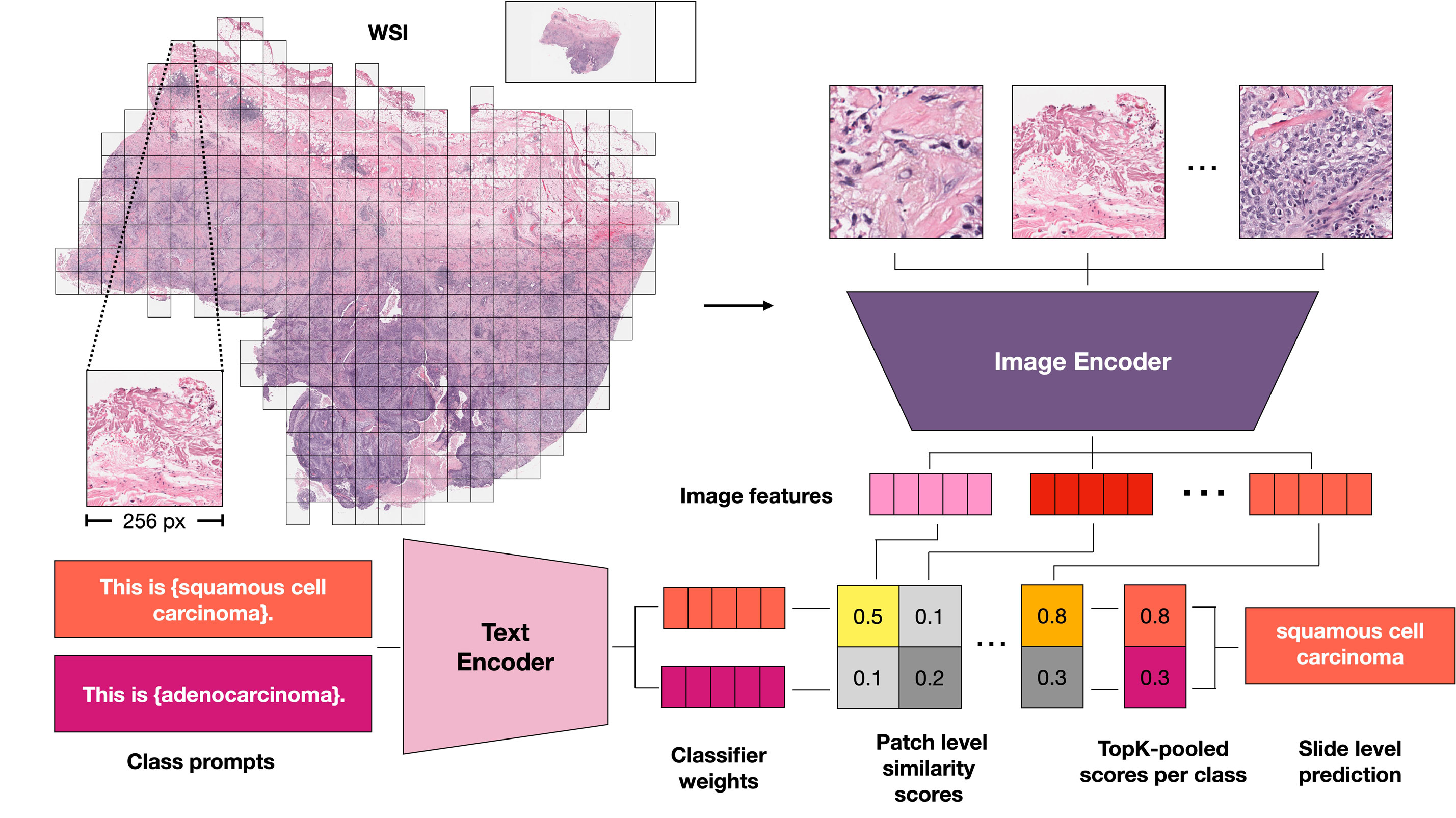}

\caption{\textbf{Schematic of MI-Zero.} A gigapixel WSI is converted to a collection of patches (instances), each embedded into an aligned visual-language latent space. In the set-based representation, the similarity scores between patch embeddings and prompt embeddings are aggregated via a permutation invariant operator such as topK max-pooling to produce the WSI-level classification prediction. Alternatively, a graph-based representation may be used to incorporate spatial context by first aggregating predictions in local neighborhoods (\Cref{sec:methods}).}
   \label{fig:wsi}
\end{figure*}

\vspace{-0.3cm}
\subsection{Zero-shot transfer for gigapixel WSIs.}
\label{sec:methods}
There are several key challenges in performing zero-shot transfer for WSIs in the manner described in the previous section. First, each WSI can span up to $100,000 \times 100,000$ pixels, making it computationally intractable to directly compute an embedding vector using the image encoder~\cite{campanella2019clinical}. Second, WSIs are known to be heterogeneous, comprising various tissue and cell types that can interact to form higher level architectures of both normal and diseased morphological patterns~\cite{heindl2015mapping, javed2020cellular, hosseini2019atlas, abousamra2021multi}. In light of these challenges we propose MI-Zero, a zero-shot transfer framework for classifying WSIs inspired by the success of multiple instance learning for solving weakly-supervised learning tasks in computational pathology.

The approach entails first dividing each WSI into smaller tiles (called instances) more amenable to processing via our image encoder. We then consider the WSI as a collection of such instances by adopting either a permutation invariant set-based representation or a graph-based representation. Specifically, we consider dividing the tissue region of each WSI into $N$ patches, and compute the $\ell_2$-normalized embeddings of each patch independently using the image encoder to obtain $\{\mathbf{u}_i\}_{i=1,\ldots,N}$. We note that $N$ is not a fixed constant, but instead varies depending on how large each WSI is, and therefore how many patches are obtained. Following the aforementioned prompt-based classification approach, we compute scores $\{\mathbf{s}_i\}_{i=1,\ldots,N}$:
\begin{equation}
\mathbf{s}_i = \mathbf{u}_i^T [\mathbf{w}_1, \mathbf{w}_2, \ldots, \mathbf{w}_C], \quad \mathbf{s}_i \in \mathbb{R}^{C}
\end{equation}
by measuring the cosine similarity between each patch embedding $\mathbf{u}_i$ and prompt embeddings $\{\mathbf{w}_m\}_{m=1,\ldots, C}$ defined earlier. 

In the set-based representation, the set of scores $\mathcal{S} = \{\mathbf{s}_i\}_{i=1,\ldots,N}$ is directly passed to any permutation invariant operator $h(\cdot)$ such as the mean operator $h_\mathrm{mean}$ or topK max-pooling operator $h_\mathrm{topK}$ to produce the slide-level prediction scores (illustrated in \Cref{fig:wsi}):
\begin{equation}
    h_\mathrm{mean}(\mathcal{S}) = \frac{1}{N}\sum_{i=1}^{N} \mathbf{s}_i
\end{equation} 
\begin{equation}
h_\mathrm{topK}(\mathcal{S}) = \frac{1}{K}\left[\sum_{i=1}^{K} \tilde{s}^{1}_i, \sum_{i=1}^{K} \tilde{s}^{2}_i, \ldots, \sum_{i=1}^{K} \tilde{s}^{C}_i\right]^T 
\end{equation} 
where $\mathcal{S}_{\mathrm{topK}}^{c} = \{\tilde{s}^c_i\}_{i=1,\ldots,K}$ is the set of the $K$ largest score values from $\mathcal{S}$ for class $c = 1, \ldots, C$. We note that in principle any permutation invariant pooling operator may be used here, as long as it is free of any learnable parameters (which are required in many attention-based methods) given the goal is to perform zero-shot transfer and no parameter update is allowed. 
In the graph-based representation, we take into account the spatial positions of each patch and build a directed KNN (k-nearest neighbors) graph $G = \{\mathcal{M}, \mathcal{E}\}$ connecting each patch (node) to its spatial neighbors, where the value at node $i$ is its scores $\mathbf{s}_i$. Given the graph representation, we spatially smooth (\textit{e.g.} average) the score values, by replacing $\mathbf{s}_i$ with $h_\mathrm{mean}(\mathcal{S}_\mathrm{neighbors})$, where $\mathcal{S}_\mathrm{neighbors} = \{\mathbf{s}_j: j \in \{i\} \cup \mathcal{N}(i)\}$ and $\mathcal{N}(i) = \{j: (i, j) \in \mathcal{E}\}$ for each node $i$ in the graph. We note that this is equivalent to applying a mean-filter with the receptive field size covering each patch's k-nearest neighbors. In principle, other filters such as the median or Gaussian filter can also be used, although we choose the simplest option (\textit{i.e.} the mean) and leave other options for future exploration. After spatial smoothing, we then proceed by applying one of the possible permutation invariant pooling operators to the set of smoothed scores in the graph, $\mathcal{S}_{\mathrm{smoothed}}$, and arrive at the slide-level prediction scores. 


\begin{figure*}
  \centering
   \includegraphics[width=1\linewidth]{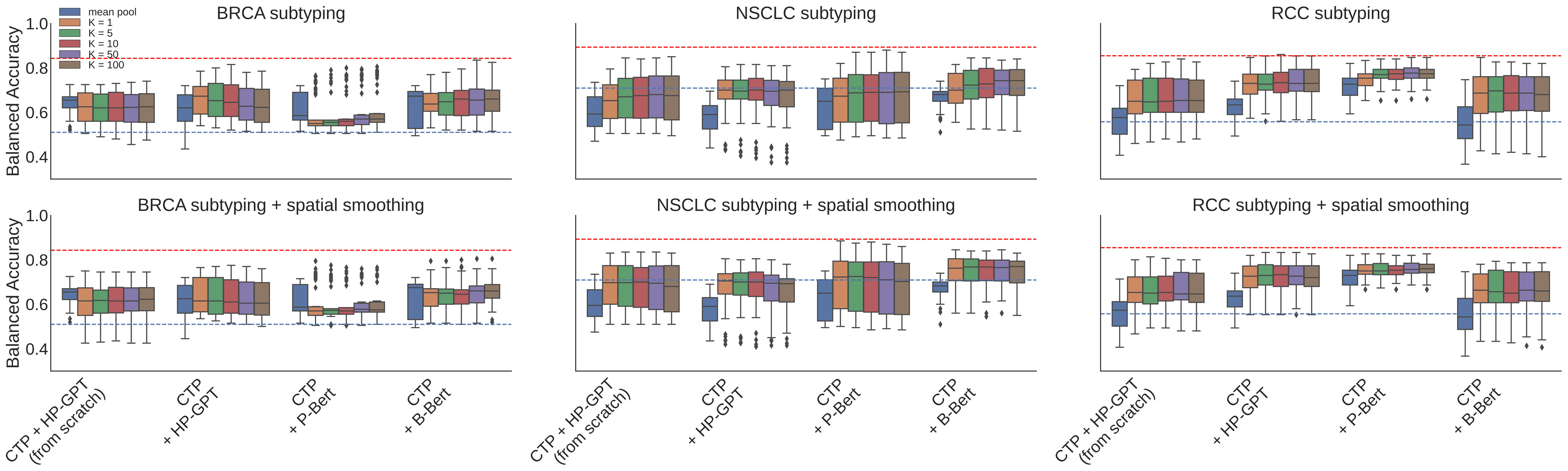}
   \vspace{-6mm}
   \caption{\textbf{Zero-shot transfer} performance of selected model configurations on independent test sets. Boxplots show distribution of model performance for 50 randomly sampled prompts. Columns show different subtyping tasks, rows show the absence or presence of spatial smoothing before pooling, and colors within each boxplot group show pooling methods ($K$ indicates the number of patches selected by topK pooling). Red dashed line shows balanced accuracy of ABMIL trained on 100\% of the corresponding TCGA cancer subsets averaged across 5 folds. Blue dashed line shows ABMIL performance trained on 1\% of training data instead. \textbf{HP-GPT}: HistoPathGPT, \textbf{P-Bert}: PubMedBert, \textbf{B-Bert}: BioClinicalBert.}
   \label{fig:boxplots}
\end{figure*}

\section{Experiments and results}
\label{sec:experiments}
\subsection{Visual language pretraining.}
For our HistPathGPT, we use a custom tokenizer trained on our dataset using Byte Pair Encoding (BPE) with a vocabulary size of 32,000. We use the hidden state corresponding to the position of the last word token to model the text embedding. For BioClinicalBert and PubmedBert we use the publicly available model weights and tokenizers as provided and use the [CLS] token as the text embedding. 
For the image encoder, by default we consider the SOTA CTransPath\cite{wang2022transformer} (CTP) encoder trained using self-supervised representation learning on unlabeled histopathology patches, which has been shown to outperform ImageNet-initialized features by a wide margin on a range of different downstream tasks in CPATH. 
For all models, we use a linear projection head to map both the text and image embeddings into a 512-dimensional latent space for alignment.
We align the representation space of our image and text encoders using the cross-view contrastive loss formulated in \Cref{subsec:vl}. We first preprocess all images to be of size 448 $\times$ 448 where images too large are first resized to 448 on the short side and then center cropped, while smaller images are zero-padded if needed. We use simple data augmentation techniques including horizontal and vertical flips applied to both images and captions. Each model is trained for 50 epochs on 8 A100 GPUs with a global batch size of 512. 
Other relevant hyperparameters and details on pretraining are included in the \textbf{Supplementary Materials}.

\subsection{Downstream datasets.}
After training on the image-text pairs as described in the previous section, we evaluate the zero-shot transfer performance for cancer subtype classification on 3 WSI datasets from Brigham and Women's Hospital. We used in-house independent datasets for zero-shot transfer evaluation because SSL encoders such as CTP are often trained on large public data repositories using all data that are available. This may result in information leakage if we use subsets of these public data sources for downstream evaluation, since their distribution was already exposed to the SSL encoder during its pretraining (transductive learning). Our in-house datasets are summarized below:

\textbf{Independent BRCA} is a dataset of invasive breast carcinoma (BRCA) histopathology WSIs. It consists of 100 slides of invasive ductal carcinoma (IDC) and 100 slides of invasive lobular carcinoma (ILC). 

\textbf{Independent NSCLC} is a dataset of non-small cell lung cancer (NSCLC) histopathology WSIs. It consists of 100 slides of lung adenocarcinoma (LUAD) and 100 slides of lung squamous cell carcinoma (LUSC).

\textbf{Independent RCC} is a dataset of renal cell carcinoma (RCC) histopathology WSIs. It consists of 50 slides of chromophobe renal cell carcinoma (CHRCC), 50 slides of clear-cell renal cell carcinoma (CCRCC), 50 slides of papillary renal cell carcinoma (PRCC). 

We include results on the publicly available datasets from The Cancer Genome Atlas (TCGA)\footnote{\href{https://portal.gdc.cancer.gov/}{portal.gdc.cancer.gov}} for the same 3 subtyping tasks in the \textbf{Supplementary Materials}.

\begin{table*}
  \centering
  \begin{tabular}{@{}p{3cm}|l|c|l||ccc|c}
    \toprule
    Model & Text Encoder \& Pretraining & SS & Pooling & BRCA & NSCLC & RCC & Average \\
    \midrule
    ABMIL (1\% Data) & None &  \xmark & attention & 0.510 & 0.709 & 0.557 & 0.592 \\
    ABMIL (100\% Data) & None & \xmark & attention & 0.843 & 0.893 & 0.855 & 0.864 \\
    \midrule
    \multirow{4}{*}{MI-Zero (Ours)} & HistPathGPT (None) & \xmark & topK & 0.625 & 0.680 & 0.653 & 0.653 \\
    & HistPathGPT (In-domain) & \xmark & topK & \textbf{0.673} & 0.700 & 0.733 & \textbf{0.702} \\
    & PubmedBert (Out-of-domain) & \xmark & topK          & 0.570 & 0.693 & \textbf{0.777} & 0.680 \\
    & BioclinicalBert (Out-of-domain) & \xmark & topK      & 0.660 & \textbf{0.742} & 0.697 & 0.700 \\
    \midrule
    \multirow{4}{*}{MI-Zero (Ours)} & HistPathGPT (None) & \cmark & topK & 0.623 & 0.700 & 0.653 & 0.659 \\
    & HistPathGPT (In-domain) & \cmark & topK  & 0.615 & 0.705 & 0.733 & 0.684 \\
    & PubmedBert (Out-of-domain) & \cmark & topK         & 0.577 & 0.725 & \textbf{0.760} & 0.688 \\
    & BioclinicalBert (Out-of-domain) & \cmark & topK       & \textbf{0.660} & \textbf{0.770} & 0.663 & \textbf{0.698} \\
    \midrule
    \multirow{4}{*}{MI-Zero (Ours)} & HistPathGPT (None) & \xmark & mean & 0.655 & 0.593 & 0.577 & 0.608 \\
    & HistPathGPT (In-domain) & \xmark & mean & 0.620 & 0.590 & 0.633 & 0.614 \\
    & PubmedBert (Out-of-domain) & \xmark & mean  & 0.585 & 0.650 & \textbf{0.727} & \textbf{0.654} \\
    & BioclinicalBert (Out-of-domain) & \xmark & mean   & \textbf{0.672} & \textbf{0.680} & 0.543 & 0.632 \\
    \midrule
    \multirow{4}{*}{MI-Zero (Ours)} & HistPathGPT (None) & \cmark & mean & 0.655 & 0.595 & 0.573 & 0.608\\
    & HistPathGPT (In-domain) & \cmark & mean & 0.625 & 0.590 & \textbf{0.637} & 0.617 \\
    & PubmedBert (Out-of-domain) & \cmark & mean & 0.587 & 0.650 & 0.730 & \textbf{0.656} \\
    & BioclinicalBert (Out-of-domain) & \cmark & mean & \textbf{0.675} & \textbf{0.682} & 0.543 & 0.634\\
    \bottomrule
  \end{tabular}
  \caption{\textbf{Slide-level zero-shot transfer.} All models shown here (including the supervised baseline ABMIL) use CTP as the image encoder.  For MI-Zero, in-domain pretraining refers to pretraining on a corpus of pathology-specific text we collected while out-of-domain pretraining refers to non-pathology-specific corpora (See \Cref{subsec:textpretraining}). SS means that spatial smoothing is used before pooling while topK and mean pooling refers to the pooling operator (\Cref{sec:methods}). For each task, we report the median balanced accuracy across 50 sampled sets. For topK pooling, we report the highest performance across all $K\in\{1,5,10,50,100\}$. See \Cref{fig:boxplots} for full distributions of results.}
  \label{tab:main}
\end{table*}

\setlength{\belowcaptionskip}{-15pt}
\begin{table}
  \centering
  \begin{tabular}{@{}p{2cm}|rrr|r@{}}
    \toprule
    Dataset & BRCA & NSCLC & RCC & Average \\
    \midrule
    CLIP\cite{clip} & 0.500 & 0.500 & 0.333 & 0.444 \\
    \midrule
    ARCH\cite{gamper2021multiple} & 0.625 & 0.593 & 0.540 & 0.586 
    \\
     Ours & \textbf{0.672} & \textbf{0.700} & \textbf{0.733} & \textbf{0.702}
    \\
    \bottomrule
  \end{tabular}
  \caption{\textbf{Training data comparison}. We report balanced accuracy and only show results using topK pooling with no spatial smoothing. Since spatial smoothing yields the same trend, they are included in the \textbf{Supplementary Materials}. With the same MI-Zero setup, OpenAI's CLIP model \cite{clip} trained on 400M generic image-text pairs performs no better than random chance across all tasks. To assess the added value of our image-text pairs, we trained our best performing model configuration from \Cref{tab:main} (CTP + HistPathGPT) on our full training dataset and compared to training only on ARCH (7,562 pathology pairs) \cite{gamper2021multiple}, which is a subset of our training data (33,480 pathology pairs).}
  \label{tab:arch}
\end{table}

\setlength{\belowcaptionskip}{-10pt}
\begin{table*}
  \centering
  \begin{tabular}{@{}l|l|l|l|rrr|r@{}}
    \toprule
    Image Encoder & Text Encoder & Image Pretraining & Text Pretraining & BRCA & NSCLC & RCC & Average \\
    \midrule
    CTP & HistPathGPT & SSL & In-domain & \textbf{0.672} & \textbf{0.700} & \textbf{0.733} & \textbf{0.702} \\
    ViT-S & HistPathGPT & SSL & In-domain & 0.617 & 0.625 & 0.673 & 0.639 \\
    ViT-S & HistPathGPT & ImageNet & In-domain & 0.660 & 0.525 & 0.600 & 0.595 \\
    CTP & HistPathGPT & None & None & 0.535 & 0.520 & 0.297 & 0.451 \\
    ViT-S & HistPathGPT & None & None & 0.500 & 0.510 & 0.290 & 0.433 \\
    \bottomrule
  \end{tabular}
  \caption{\textbf{Pretraining comparison}. To assess the benefit of pretraining the image encoder, we compare our best performing model with a variation that uses ViT-S pretrained using SSL (MoCo v3), pretrained using supervised ImageNet, as well as variations with entirely randomly-initialized weights. We report balanced accuracy and we only show results using topK pooling with no spatial smoothing. Since spatial smoothing yields the same trend, we include those results in the \textbf{Supplementary Materials}.}
  \label{tab:rand}
\end{table*}

\subsection{Supervised baselines}
To help contextualize the performance of zero-shot transfer models, we train supervised baselines using weakly-supervised attention-based MIL (ABMIL) \cite{ilse2018attention} on the publicly available TCGA cohort of each task. Due to the relatively small size of these datasets ($\sim$1000 WSIs each), we follow the study design of other weakly-supervised classification studies by performing 5-fold Monte Carlo cross-validation to train 5 different models and report their average performance on our in-house datasets. Each cross-validation training set includes on average 836 slides for BRCA, 838 for NSCLC, and 739 for RCC. We additionally study the data efficiency of ABMIL by restricting the number of training labels to be 1\% and 10\% of the full training set.  We include more details about the training and results in the \textbf{Supplementary Materials}.

\subsection{Zero-shot transfer}
\noindent
\textbf{Zero-shot evaluation methodology.} 
Due to the reliance on prompts for zero-shot transfer, evaluation results vary with the choice of class names and prompt templates. For each task, we first curate a pool of relevant prompt templates and classnames (see \textbf{Supplementary Materials}). We then evaluate each model configuration on each task by randomly sampling 50 prompts and measuring the performance of each prompt. We plot the accuracy achieved using each prompt and report the median and interquartile range similar to \cite{sanh2022multitask}. This provides a more holistic view of the model’s performance and demonstrates the degree of variation from using different prompts. 





\begin{figure}[t]
  \centering
   \includegraphics[width=1\linewidth]{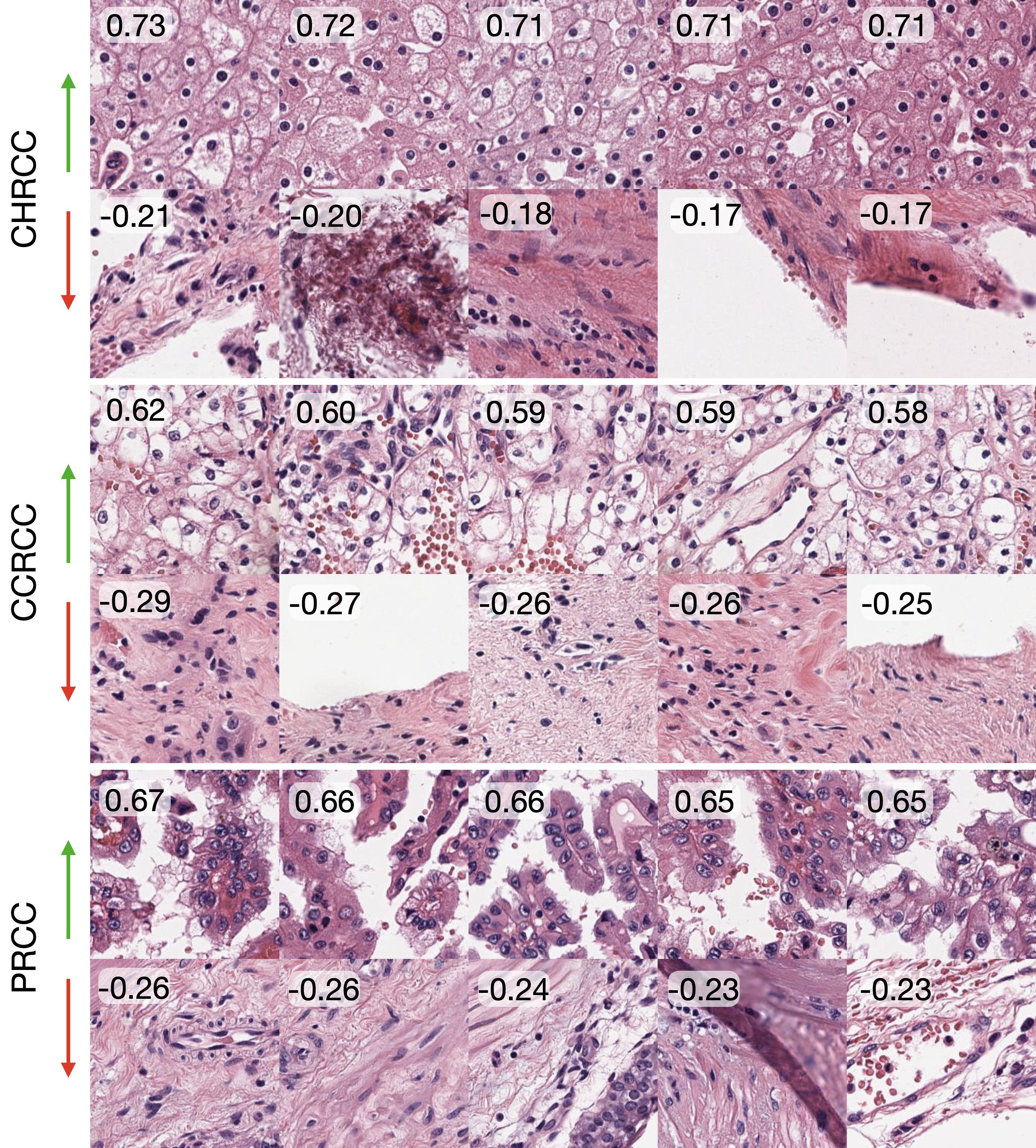}
   \caption{\textbf{Visualization of similarity scores.} A WSI of each RCC subtype (CHRCC, CCRCC and PRCC) is randomly selected from the in-house test set, and patches are ranked by their cosine similarity score with the class prompt embedding. The top (highest similarity scores) and bottom (lowest similarity scores) patches are displayed for each WSI. A board certified pathologist confirms relevant morphological patterns to each class embedding are selected by MI-Zero (high similarity scores), while low scores generally correspond to debris or normal tissue irrelevant to diagnosis. See \textbf{Supplementary Materials} for examples from other tasks.}
   \label{fig:vis}
\end{figure}

\vspace{10pt}
\noindent
\textbf{Zero-shot transfer for WSIs.} For each of the 3 cancer subtyping classification tasks, we preprocess the WSIs by segmenting the tissue regions and dividing them into 256 $\times$ 256-sized patches at the 20$\times$ equivalent magnification. We then treat each WSI as a collection of its patches (instances) similar to MIL and use MI-Zero as described in \Cref{sec:methods} for zero-shot transfer. 
We compared performance between using a text encoder pretrained on in-domain pathology text data (HistPathGPT), encoders pretrained on non-pathology-specific medical data (PubMedBert and BioClinicalBert), as well as a text encoder trained from scratch. We also experimented with different pooling methods and spatial smoothing. Classification results comparing these setups are summarized in \Cref{tab:main} and boxplots showing the performance distribution of each model on the set of 50 sampled prompts can be found in \Cref{fig:boxplots}. Overall, our models either perform on par or better than ABMIL baselines using 1\% of training data for every task. In terms of pooling method for MI-Zero, we find that topK pooling performs better than mean pooling, while spatial smoothing does not change the results significantly. We find that pretraining the text encoder improves performance over no pretraining, but pretraining on in-domain pathology text does not necessarily yield better performance. Example patches of highest and lowest similarity scores are visualized in \Cref{fig:vis}.

\setlength{\belowcaptionskip}{-15pt}

\subsection{Ablation study}
\noindent
\textbf{Training data comparison.} To assess the benefit of pretraining with our expanded image-text dataset (compared to the smaller publicly available ARCH dataset originally proposed for representation learning via captioning), we train our best performing model configuration (CTP as image encoder and HistPathGPT pretrained on in-domain data as text encoder) on ARCH only. 
We find that training on our larger dataset improves performance across all tasks and raises the overall average performance by 11.6\%  (\Cref{tab:arch}).

\vspace{10pt}
\noindent
\textbf{Image encoder pretraining.} We experimented with the choice of image encoder by comparing CTP to encoders based on the ViT-S architecture, which has a similar parameter count. The encoders evaluated include both ImageNet initialization and pretraining with SSL (MoCo v3\cite{chen2021empirical}) on in-domain histology image data\cite{wang2022transformer}. We also evaluate both the CTP and ViT-S encoders initialized fully from scratch with no pretraining as an additional ablation study. We find that pretraining both the image encoder and the text encoder performs the best across all tasks (\Cref{tab:rand}).


\vspace{3mm}
\noindent
\textbf{Locked-image tuning.} Zhai \textit{et al.}\cite{lit} recently showed that ``locking" a well-pretrained image encoder outperforms its unlocked counterpart during contrastive tuning. We therefore also explored locked-image tuning by freezing the parameters in the pretrained image encoder and only updating the text encoder. We find that when using the SSL-pretrained CTP as the image encoder and in-domain pretrained HistPathGPT as the text encoder, locked-image text tuning only offers marginal improvement on zero-shot transfer performance. For all other configurations, 
locked-image tuning considerably lowers performance. 
We conjecture that by pretraining on in-domain data, the image and text features are easier to align in the latent space such that locked-image tuning was able to provide an improvement (See \textbf{Supplementary Materials)}.

\vspace{3mm}
\noindent
\textbf{Additional experiments.} Additional experimental results using TCGA WSIs, as well as run time analyses of MI-Zero, are included in \textbf{Supplementary Materials}.

\vspace{-0.5mm}
\section{Conclusion}
\label{sec:conclusion}

In this work we introduce MI-Zero, the first method for zero-shot transfer in pathology, and apply it to gigapixel-scale whole slide images. The ability of our visual language pretrained model to retrieve relevant ROIs for a given class label (see \Cref{fig:vis} with additional examples in \textbf{Supplementary Materials}) suggests potential usefulness for semi-supervised learning workflows in histopathology\cite{bulten2020automated, elforaici2022semi, srinidhi2022self, qu2022towards} (e.g. as a way of performing pseudo-labeling). Our current results, however, are constrained by data limitations, as curating larger datasets of high quality image-caption pairs is a difficult task. Valuable future directions include collecting additional image caption datasets\cite{tsuneki2022inference, gamper2021multiple}, exploring methods that may improve the sample efficiency of visual language pretraining and also evaluating the capabilities of zero-shot transfer models on a large and diverse set of computational pathology benchmarks. We hope our work might inspire efforts to curate large scale pathology-specific image text datasets, and pave the way for a new generation of models in computational pathology capable of performing diverse visual language understanding tasks such as visual question answering, cross-modal retrieval, and captioning. Lastly, beyond pathology, many fields including satellite imaging and remote sensing involve high resolution images, similar to WSIs, in their workflow. MI-Zero can potentially be generalized and adapted to create effective solutions in such domains. 


\section{Acknowledgements}
We thank Guillaume Jaume for his feedback. This work was supported in part by the BWH president’s fund, BWH \& MGH Pathology, and NIGMS R35GM138216 (F.M.). M.Y.L. was also supported by the Siebel Scholars program. R.J.C. was also supported by the NSF Graduate Fellowship. 

{\small
\bibliographystyle{ieee_fullname}
\bibliography{egbib}
}

\end{document}


\twocolumn[
\centering

\Large
\textbf{Supplementary Material} \\
\vspace{1.0em}
] 
\appendix

\section{Additional dataset details}
\subsection{Image-caption dataset preprocessing}
Our cleaning and filtering workflow consist of first filtering out invalid image-caption pairs (e.g. either if the caption is empty/nonsensical, or if the corresponding image is missing). We then, to the best of our knowledge, removed non-histopathology images (including gross images, cytology images, X-ray/CT images, electron microscopy images, fluorescent microscopy images, schematic diagrams, etc.) and cropped multipanel figures into individual images, cleaning the paired caption accordingly. The resulting dataset is highly diverse and is expected to cover all major biopsy sites and morphologies of both diseased and normal tissue. 

\begin{figure}[h]
  \centering
   \includegraphics[width=0.8\linewidth]{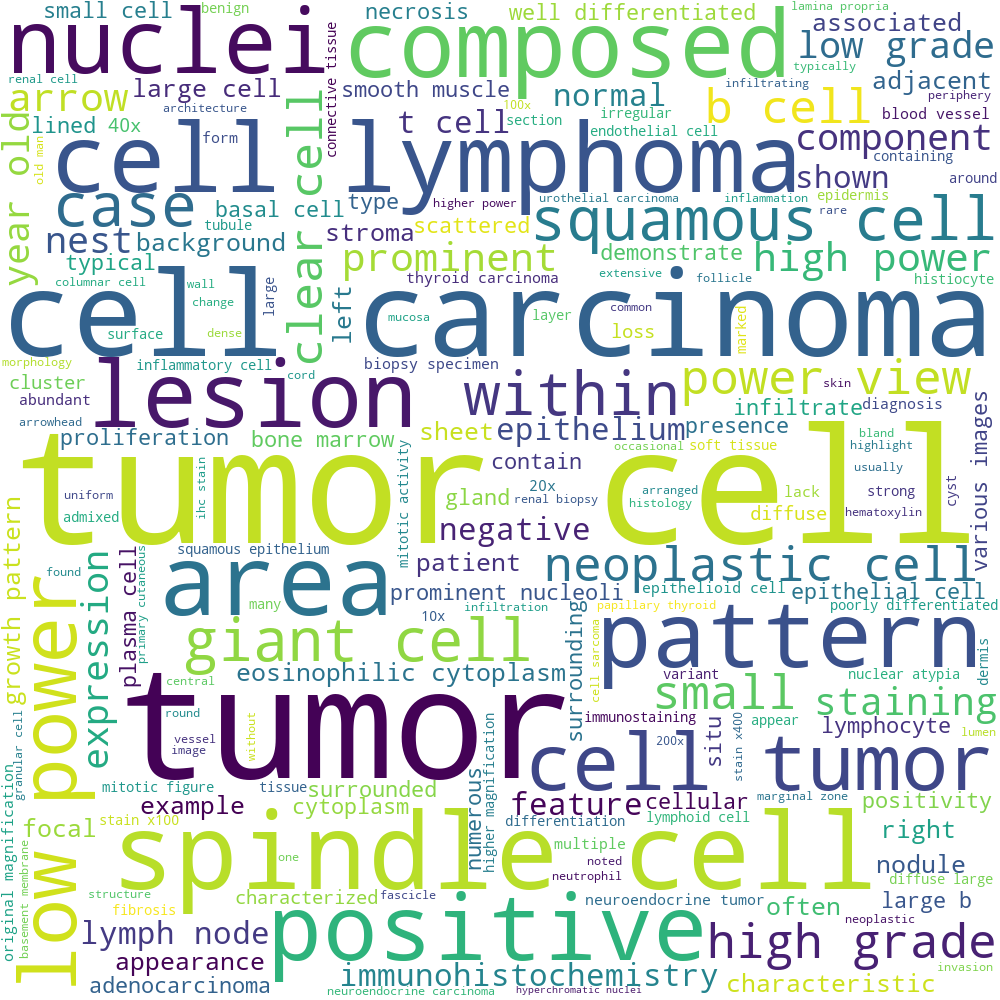}
   \caption{\textbf{Word cloud of captions in the paired image-caption dataset used for pretraining.} For clarity, we excluded some common verbs and combined spelling variations of the same word.}
   \label{fig:compare_op}
\end{figure}

\subsection{Color variation and diversity of images} 
\begin{figure}[h]
  \centering
   \includegraphics[width=1.\linewidth]{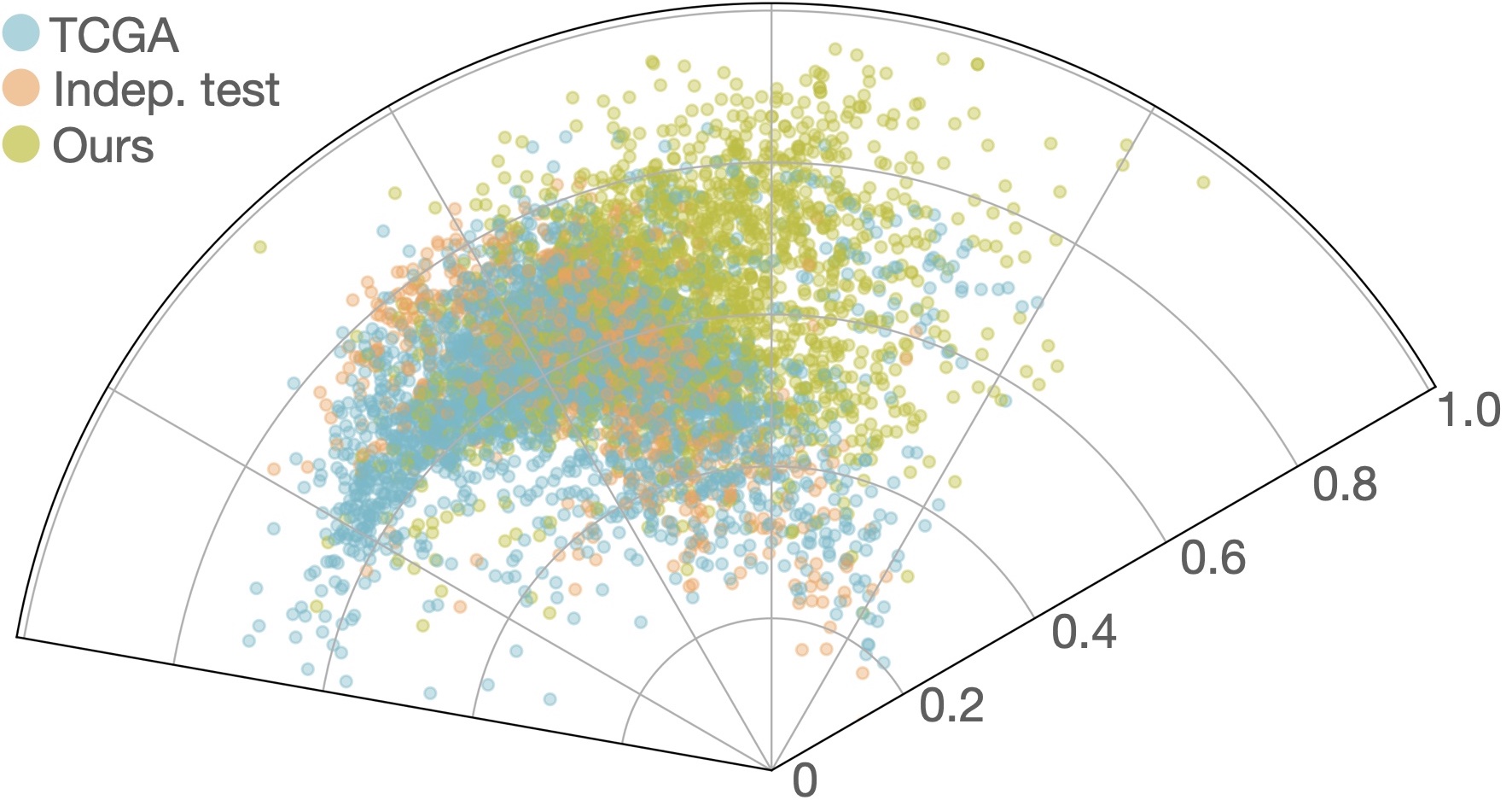}
   \vspace{-3mm}
   \caption{Plot of mean hue (angle) and saturation (radius) of each image (dot) after RGB to Hue-Saturation-Density (HSD) transformation. Sampling was performed to avoid over clutter.}
   \label{fig:color}
\end{figure}

Recent large scale studies have demonstrated that stain normalization (SN) is not required to achieve robust generalization \cite{tellez2019quantifying}
, especially when training with large, diverse, multi-institutional datasets that cover a wide range of staining profiles \cite{campanella2019clinical}. Therefore we did not perform SN to avoid computational overhead and choosing between SN algorithms. We investigated staining variations in the datasets used (Fig. \ref{fig:color}). We find TCGA and our dataset show wider coverage due to sourcing from many institutions and/or diverse tissue sites compared to our in-house independent test set, but we observe substantial overlap across datasets.
\setlength{\belowcaptionskip}{-14pt}
\vspace{-3mm}

\section{Additional training details}
\subsection{HistPathGPT pretraining}
See \Cref{tab:hparams_gpt}.
\begin{table}[!htbp]
  \centering
  \begin{tabular}{@{}p{3.5cm}|rrr|r@{}}
    \toprule
    Hyperparameter & Value \\
    \midrule
    Architecture & gpt2-medium \\
    Max. sequence length & 512 \\
    Vocabulary size & 32000 \\
    \midrule
    Batch size & 64 \\
    Gradient accumulation & 4 \\
    Weight decay & 0.01 \\
    AdamW $\beta$ & (0.9, 0.999) \\
    Peak learnng rate & 1e-4 \\
    Learning rate schedule & Linear \\
    Warmup steps & 750 \\
    Training steps & 15000 \\
    \bottomrule
  \end{tabular}
  \caption{\textbf{Hyperparameters used in pretraining the text encoder (HistPathGPT)}. In-house pathology reports were first de-identified using regex pattern matching before tokenization. 4 $\times$ 80GB NVIDIA A100 GPUs were used for training. Batch size refers to the total batch size across GPUs. Effective batch size used for optimization is batch size $\times$ gradient accumulation steps. The sequence length of training examples was set to the maximum sequence length supported by the model (\textit{i.e.} 512).}
  \label{tab:hparams_gpt}
\end{table}
\subsection{Visual language pretraining}
\label{app:vlpt}
See \Cref{tab:hparams_pretrain}.
\begin{table}[!htbp]
  \centering
  \begin{tabular}{@{}p{3.5cm}|rrr|r@{}}
    \toprule
    Hyperparameter & Value \\
    \midrule
    Batch Size & 512 \\
    Weight decay & 0.2 \\
    AdamW $\beta$ & (0.9, 0.999) \\
    Temperature & 0.07 \\
    Peak learning rate & 1e-4 \\
    Learning rate schedule & Cosine \\
    Warmup steps & 100 \\
    Epochs & 50 \\
    \bottomrule
  \end{tabular}
  \caption{\textbf{Hyperparameters used in visual language pretraining}. 8 $\times$ 80GB NVIDIA A100 GPUs were used for training. The maximum sequence length for captions is set to 128.}
  \vspace{3mm}
  \label{tab:hparams_pretrain}
\end{table}

\subsection{Supervised baselines}
\label{app:supbase}
\begin{figure*}[h]
  \centering
   \includegraphics[width=1\linewidth]{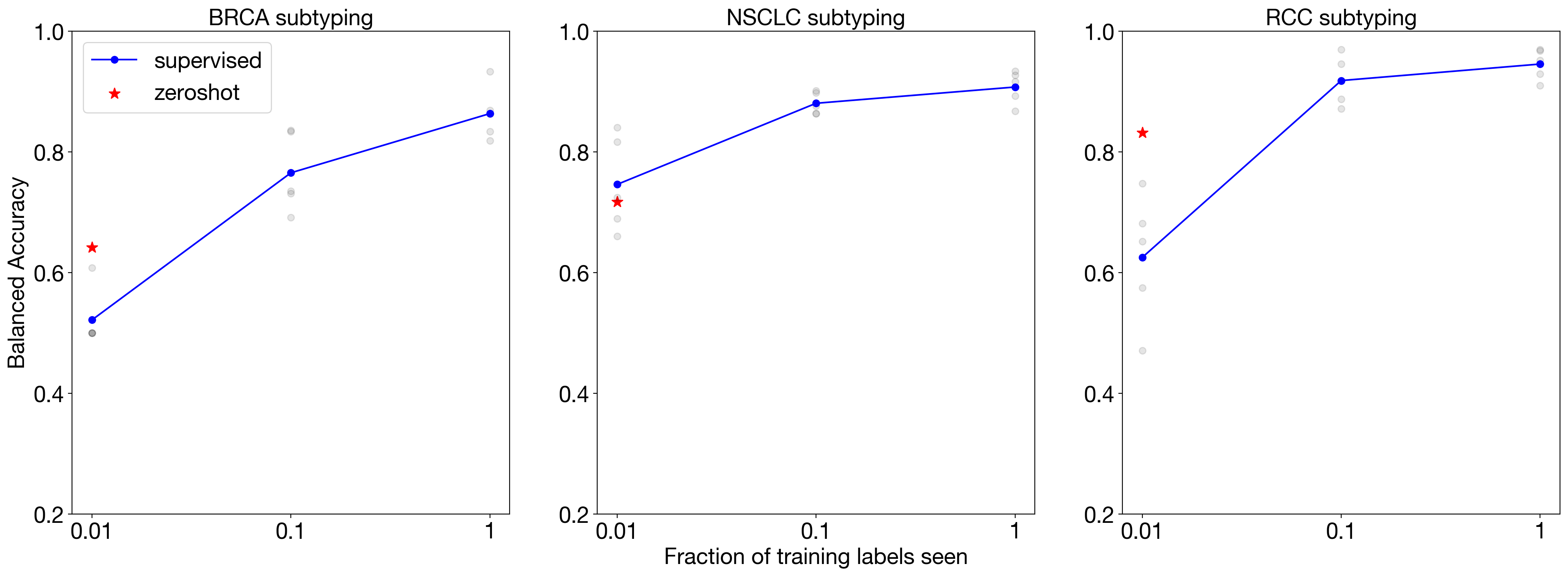}
   \caption{\textbf{ABMIL supervised baseline and MI-Zero zeroshot performance on TCGA test sets.} The topK pooling variant of MI-Zero is shown. For varying label fractions, each ABMIL model from the 5-fold CV run is represented by a gray dot and the 5-fold average is represented by the blue dot.}
   \label{fig:compare_tcga}
\end{figure*}

\begin{table}[h]
  \centering
  \begin{tabular}{@{}p{3.5cm}|rrr|r@{}}
    \toprule
    Hyperparameter & Value \\
    \midrule
    Weight decay & 1e-5 \\
    AdamW $\beta$ & (0.9, 0.999) \\
    Learning rate & 1e-4 \\
    \bottomrule
  \end{tabular}
  \caption{\textbf{Hyperparameters used in weakly supervised baselines}. Each experiment is performed on a single 24GB NVIDIA 3090 GPU.}
  \label{tab:hparams_supbase}
\end{table}
To establish a supervised baseline of comparison for zeroshot transfer, we use the ABMIL weakly-supervised learning algorithm\cite{ilse2018attention}. Same as MI-Zero, instances (\textit{i.e.} patches of size 256 $\times$ 256 at 20$\times$-equivalent magnification) are embedded using the SOTA SSL encoder CTransPath\cite{wang2022transformer} (CTP).
Due to the relatively small size of the TCGA datasets ($\sim 1000$ slides or fewer for each task), we follow the study design of previous works by performing 5-fold Monte Carlo cross-validation (CV). In each fold, each dataset is randomly partitioned at the patient level into training (80\%), validation (10\%) and testing (10\%), stratified by the class label. The validation set is used to early stop training and model selection and performance on the test set is reported (\Cref{app:tcga}). Each model is trained for a minimum of 20 epochs and an early stopping patience of 10 epochs based on the balanced accuracy measured on the validation set. The 5 models trained using 5-fold CV are then evaluated on our independent, in-house datasets described in \textbf{Section 4.2}. For experiments using 1\% and 10\% of training labels, in each of 5 folds, we perform stratified sampling to select 1\% and 10\% of the full training set as the new training sets. All results are included in \Cref{fig:compare_op} and \Cref{fig:compare_tcga}.

\section{Additional zero-shot transfer details}
\subsection{Prompt pools}
\label{app:prompts}

To evaluate zeroshot transfer, we sampled 50 sets of prompts using a predetermined prompt pool for each task. Each task draws from a unique pool of class names while the pool of templates are shared across tasks. A class name and the template collectively forms the prompt. For the templates, we have:

\begin{itemize}
    \setlength\itemsep{-5.5mm}
    \item \texttt{CLASSNAME.} \\
    \item \texttt{a photomicrograph showing CLASSNAME.} \\
    \item \texttt{a photomicrograph of CLASSNAME.} \\
    \item \texttt{an image of CLASSNAME.} \\
    \item \texttt{an image showing CLASSNAME.} \\ 
    \item \texttt{an example of CLASSNAME.} \\
    \item \texttt{CLASSNAME is shown.} \\
    \item \texttt{this is CLASSNAME.} \\
    \item \texttt{there is CLASSNAME.} \\
    \item \texttt{a histopathological image showing CLASSNAME.} \\
    \item \texttt{a histopathological image of CLASSNAME.} \\
    \item \texttt{a histopathological photograph of CLASSNAME.} \\
    \item \texttt{a histopathological photograph showing CLASSNAME.} \\
    \item \texttt{shows CLASSNAME.} \\
    \item \texttt{presence of CLASSNAME.} \\
    \item \texttt{CLASSNAME is present.}
\end{itemize}

The \texttt{CLASSNAME} is replaced by a sampled class name. The class names for each task are presented in \Cref{tab:classnames}. For each of 50 prompts, we sample a random number of templates and ensemble them in the embedding space in the same manner as performed by CLIP\cite{clip}.

\subsection{Additional results to ablation studies}
\label{app:ablations}

\noindent \textbf{Training data comparison.} To assess the added value of our image-text pairs, we trained our best performing model configuration (CTP + HistPathGPT) on our full training dataset and compared to training only on ARCH (7,562 pathology pairs), which is a subset of our training data (33,480 pathology pairs). We find that for all pooling methods, training on our full dataset performs better than training on ARCH only. This trend did not change significantly across different pooling methods with MI-Zero. Results are presented in \Cref{tab:archsupp}.

\vspace{10pt}
\noindent \textbf{Image encoder pretraining.} To assess the benefit of pretraining the image encoder, we compare our best performing model with a variation that uses ViT-S pretrained on ImageNet as well as starting with entirely randomly-initialized weights. We find that pretraining the image encoder and the text encoder on in-domain data performs the best across all 3 tasks. The same trend is maintained across different pooling methods. We report full results in \Cref{tab:randsupp}.


\vspace{10pt}\noindent \textbf{Locked-image tuning.}
We assess whether locked image tuning (LiT)\cite{lit} improves performance on zeroshot transfer by freezing all parameters in the pretrained image encoder when performing visual-language pretraining. Results are reported in  \Cref{tab:lit}. We find that LiT improves performance slightly when the text encoder is pretrained on in-domain corpora (HistPathGPT), but significantly degrades performance otherwise. We hypothesize this is because when the image and text encoders are pretrained on different domains, the domain gap between the image and text latent space makes it challenging for the model to align via a linear projection and additional parameters and non-linearity might need to be needed. 

\subsection{Visualization of similarity scores}
\label{app:simscore}

\begin{figure*}[h]
  \centering
   \includegraphics[width=1\linewidth]{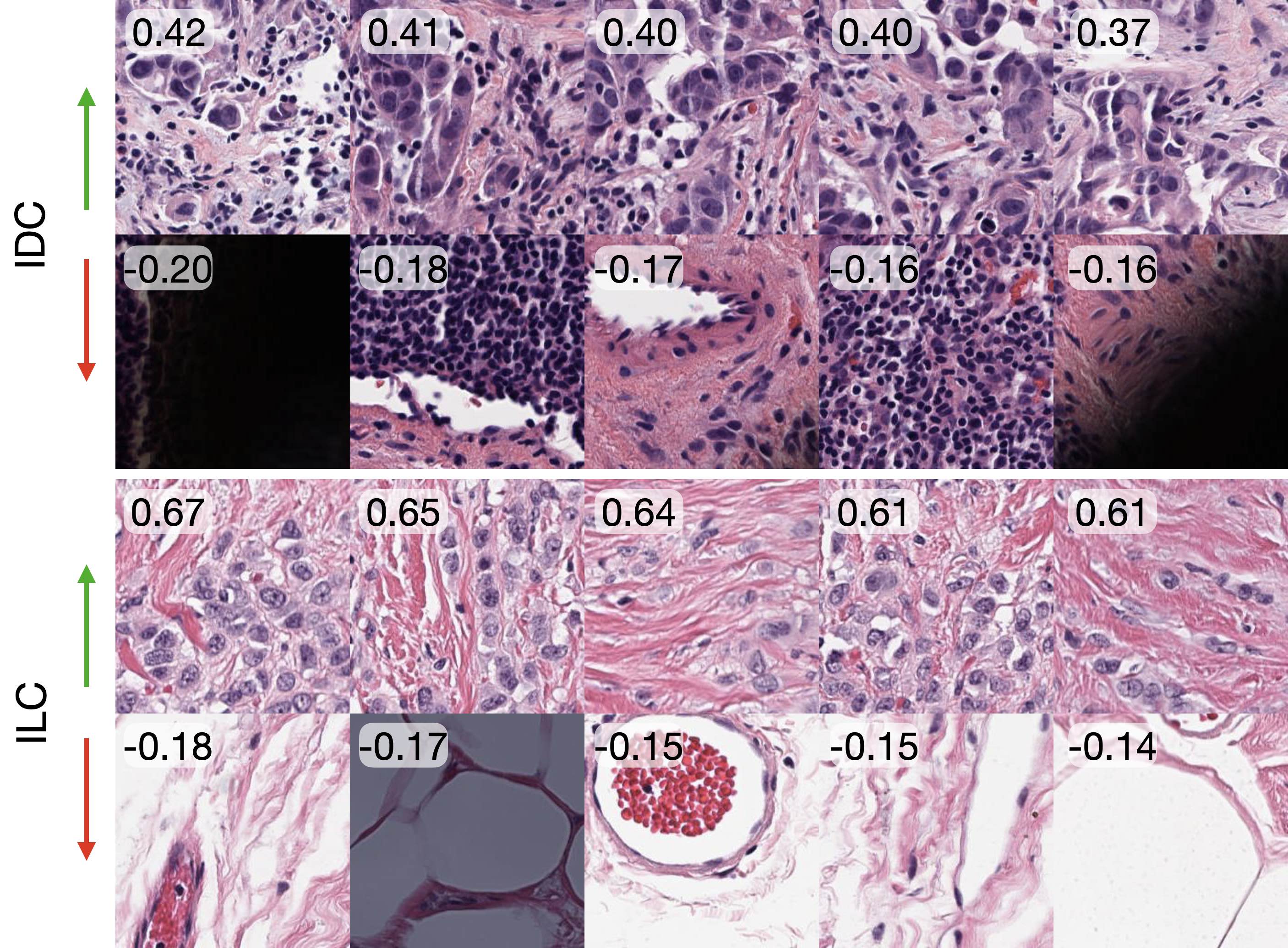}
   \caption{\textbf{Visualization of similarity scores for BRCA subtyping.} A WSI of each BRCA subtype (IDC, ILC) is randomly selected from the in-house independent test set, and patches are ranked by their cosine similarity score with the class prompt embedding. The top (highest similarity scores) and bottom (lowest similarity scores) patches are displayed for each WSI. A board-certified pathologist confirms relevant morphological patterns to each class embedding are selected by MI-Zero (high similarity scores). In IDC, high scores correspond to moderate grade tumor cells forming nests and abortive glands, typical of invasive ductal carcinoma of the breast. Low scores picked up instances covered by pen markings, connective tissue, and lymphocyte aggregates with no tumor cells present.
   In ILC, high scores correspond to low grade tumor cells invading as individual cells or as single files, typical of invasive lobular carcinoma of the breast. Low scores picked up instances of connective and adipose tissue with no tumor cells present.}
   \label{fig:vis_brca}
\end{figure*}

\begin{figure*}[h]
  \centering
   \includegraphics[width=1\linewidth]{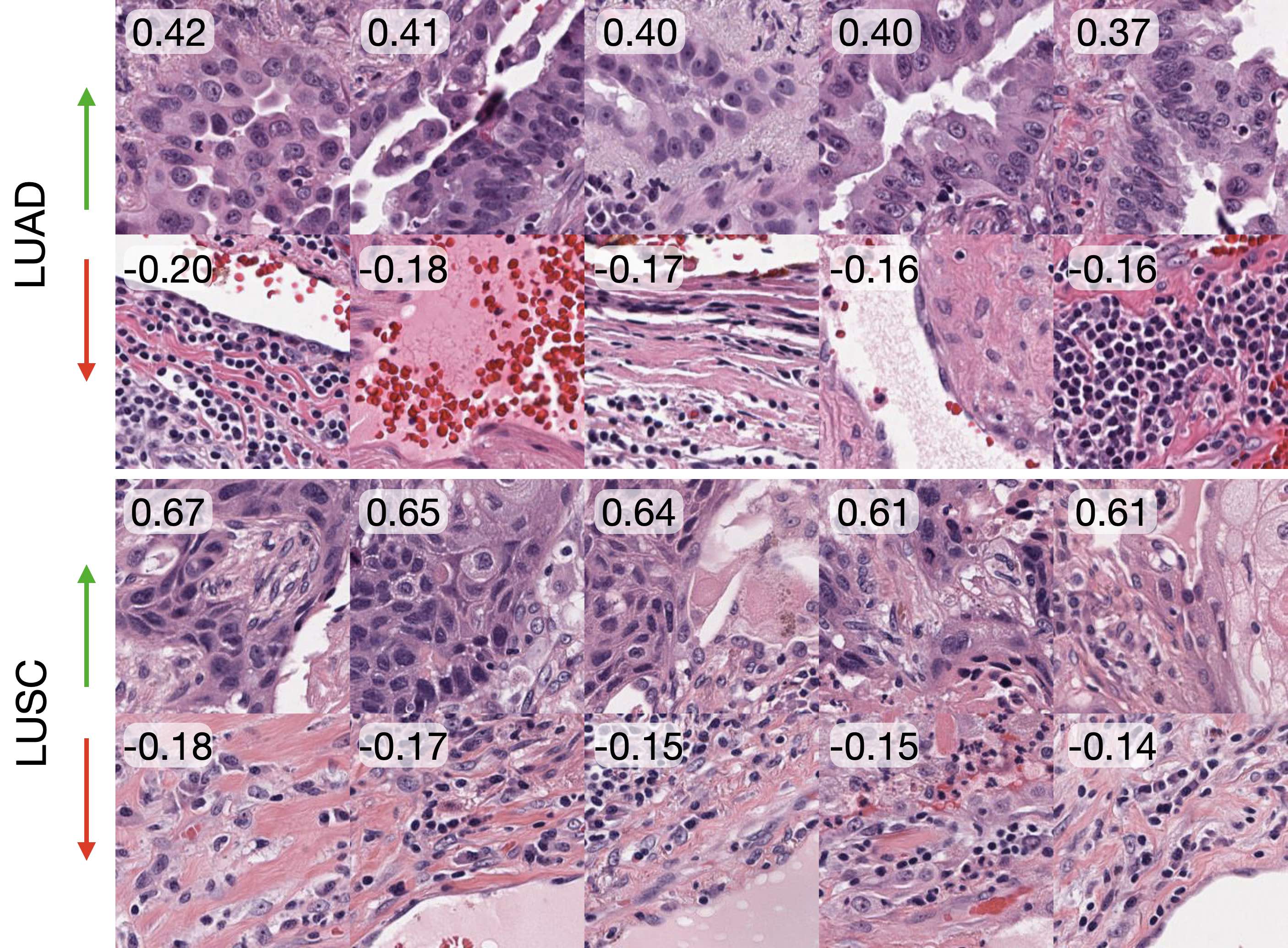}
   \caption{\textbf{Visualization of similarity scores for NSCLC subtyping.} A WSI of each NSCLC subtype (LUAD, LUSC) is randomly selected from the in-house independent test set, and patches are ranked by their cosine similarity score with the class prompt embedding. The top (highest similarity scores) and bottom (lowest similarity scores) patches are displayed for each WSI. A board-certified pathologist confirms relevant morphological patterns to each class embedding are selected by MI-Zero (high similarity scores). In LUAD, high scores correspond to tumor cells forming nests and lining spaces with prominent nucleoli, characteristic of adenocarcinoma. Low scores picked up connective tissue, lymphocytes, and blood.
   In LUSC, high scores correspond to aggregates of tumor cells with keratinization and intercellular bridges characteristic of squamous cell carcinoma. Low scores picked up connective tissue, muscle, and inflammatory cells.}
   \label{fig:vis_nsclc}
\end{figure*}

We randomly select a slide from each in-house test set and visualize patches with highest and lowest similarity scores (\Cref{fig:vis_brca} and \Cref{fig:vis_nsclc}). A board-certified pathologist confirmed that relevant morphological patterns are selected with high similarity scores, which drive the model's zero-shot slide-level predictions in the case of topK pooling. 

\subsection{Additional zero-shot transfer on TCGA}
\label{app:tcga}

\begin{figure*}
  \centering
   \includegraphics[width=1\linewidth]{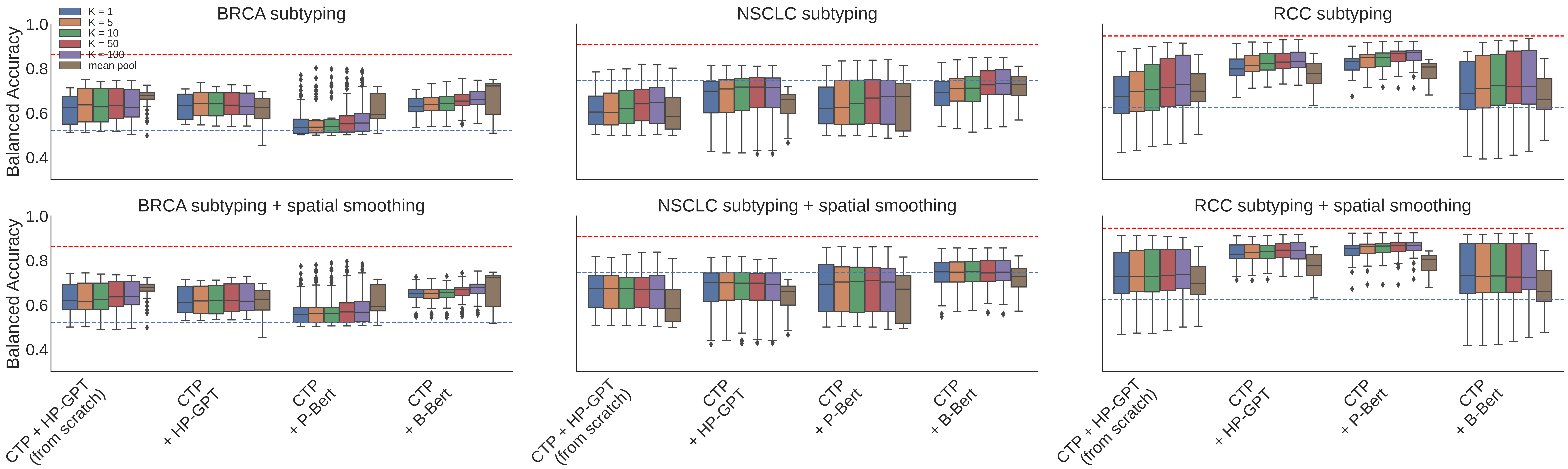}
   \vspace{-6mm}
   \caption{\textbf{Zero-shot transfer} performance of selected model configurations on TCGA subsets. Boxplots show distribution of 5-fold averaged model subsets performance for 50 randomly sampled prompts. Columns show different subtyping tasks, rows show the absence or presence of spatial smoothing before pooling, and colors within each boxplot group show pooling methods ($K$ indicates the number of patches selected by topK pooling). Red dashed line shows balanced accuracy of ABMIL trained on 100\% of the corresponding TCGA cancer subsets averaged across 5 folds. Blue dashed line shows ABMIL performance trained on 1\% of training data instead. \textbf{HP-GPT}: HistoPathGPT, \textbf{P-Bert}: PubMedBert, \textbf{B-Bert}: BioClinicalBert.}
   \label{fig:boxplots_tcga}
\end{figure*}

As introduced in the main paper, we present additional results on the TCGA counterparts of our independent test set tasks (BRCA, NSCLC, and RCC). For each of the 3 cancer subtyping classification tasks, we preprocesses the WSIs the same way as presented in the main paper and use MI-Zero for zero-shot transfer. 
Classification results comparing several setups are summarized in \Cref{tab:main_tcga} and boxplots showing the performance distribution of each model on the set of 50 sampled prompts can be found in \Cref{fig:boxplots_tcga}. Overall, we observe consistent trends between our independent test set results and TCGA results.  




\begin{figure*}[h]
  \centering
   \includegraphics[width=1\linewidth]{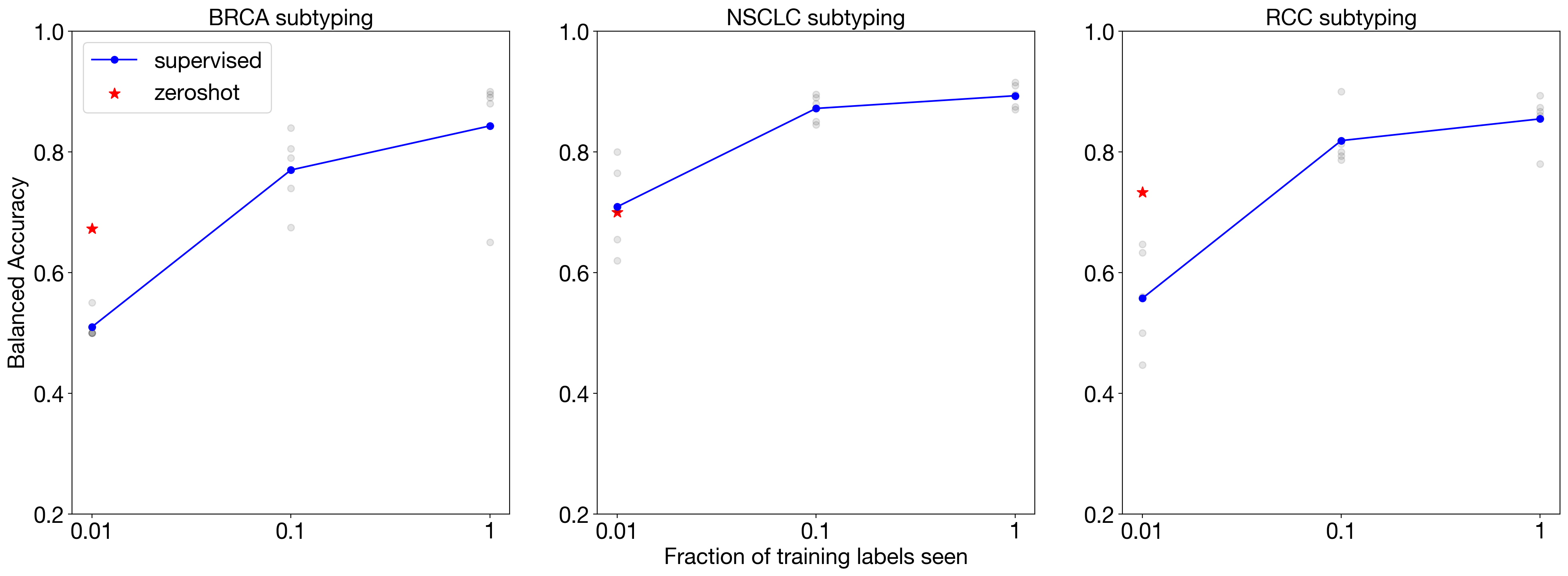}
   \caption{\textbf{ABMIL supervised baseline and MI-Zero zeroshot performance on the independent in-house datasets.} The topK pooling variant of MI-Zero is shown. For varying label fractions, each ABMIL model from the 5-fold CV run is represented by a gray dot and the 5-fold average is represented by the blue dot. }
   \label{fig:compare_op}
\end{figure*}

\begin{table*}[]
    \centering
    \begin{tabular}{l|l|l}
        \toprule
        Task & Class & Class names \\
        \midrule
        \multirow{4}{*}{BRCA} & \multirow{2}{*}{IDC} & \texttt{invasive ductal carcinoma} \\
        & & \texttt{carcinoma of the breast, ductal pattern} \\
        \cmidrule{2-3}
        & \multirow{2}{*}{ILC} & \texttt{invasive lobular carcinoma } \\
        & & \texttt{carcinoma of the breast, lobular pattern} \\
        \midrule
        \multirow{13}{*}{NSCLC} & \multirow{9}{*}{LUAD} & \texttt{adenocarcinoma } \\
        & & \texttt{lung adenocarcinoma } \\
        & & \texttt{adenocarcinoma of the lung } \\
        & & \texttt{pulmonary adenocarcinoma } \\
        & & \texttt{adenocarcinoma, lepidic pattern } \\
        & & \texttt{adenocarcinoma, solid pattern } \\
        & & \texttt{adenocarcinoma, micropapillary pattern } \\
        & & \texttt{adenocarcinoma, acinar pattern } \\
        & & \texttt{adenocarcinoma, papillary pattern} \\
        \cmidrule{2-3}
        & \multirow{4}{*}{LUSC} & \texttt{squamous cell carcinoma} \\
        & & \texttt{lung squamous cell carcinoma} \\
        & & \texttt{squamous cell carcinoma of the lung} \\
        & & \texttt{pulmonary squamous cell carcinoma} \\
        \midrule
        \multirow{12}{*}{RCC} & \multirow{4}{*}{CCRCC} & \texttt{clear cell renal cell carcinoma} \\
        & & \texttt{renal cell carcinoma, clear cell type} \\
        & & \texttt{renal cell carcinoma of the clear cell type} \\
        & & \texttt{clear cell RCC} \\
        \cmidrule{2-3}
        & \multirow{4}{*}{PRCC} & \texttt{papillary renal cell carcinoma} \\
        & & \texttt{renal cell carcinoma, papillary type} \\
        & & \texttt{renal cell carcinoma of the papillary type} \\
        & & \texttt{papillary RCC} \\
        \cmidrule{2-3}
        & \multirow{4}{*}{CHRCC} & \texttt{chromophobe renal cell carcinoma} \\
        & & \texttt{renal cell carcinoma, chromophobe type} \\
        & & \texttt{renal cell carcinoma of the chromophobe type} \\
        & & \texttt{chromophobe RCC} \\
    \bottomrule
    \end{tabular}
    \caption{\textbf{Class name pools} for each class in each task. Sampled subsets are substituted into sampled templates to form prompts.}
    \label{tab:classnames}
\end{table*}

\begin{table*}
  \centering
  \begin{tabular}{@{}l|c|l||ccc|c@{}}
    \toprule
    Dataset & SS & Pooling & BRCA & NSCLC & RCC & Average \\
    \midrule
    ARCH & \multirow{2}{*}{\xmark} & \multirow{2}{*}{topK} & 0.625 & 0.593 & 0.540 & 0.586 \\
    Ours & & & \textbf{0.672} & \textbf{0.700} & \textbf{0.733} & \textbf{0.702} \\
    \midrule
    ARCH & \multirow{2}{*}{\cmark} & \multirow{2}{*}{topK} & \textbf{0.635} & 0.607 & 0.523 & 0.589 \\
    Ours & & & 0.615 & \textbf{0.705} & \textbf{0.733} & \textbf{0.684} \\
    \midrule
    ARCH & \multirow{2}{*}{\xmark} & \multirow{2}{*}{mean} & \textbf{0.655} & 0.515 & 0.533 & 0.568 \\
    Ours & & & 0.620 & \textbf{0.590} & \textbf{0.633} & \textbf{0.614} \\
    \midrule
    ARCH & \multirow{2}{*}{\cmark} & \multirow{2}{*}{mean} & \textbf{0.650} & 0.518 & 0.530 & 0.566 \\
    Ours & & & 0.625 & \textbf{0.590} & \textbf{0.637} & \textbf{0.617} \\
    \bottomrule
  \end{tabular}
  \caption{\textbf{Training data comparison}. Balanced accuracies on in-house independent test sets are shown. To assess the added value of our image-text pairs, we trained our best performing model configuration (CTP + HistPathGPT) on our full training dataset and compared to training only on ARCH (7,562 pathology pairs), which is a subset of our training data (33,480 pathology pairs).}
  \label{tab:archsupp}
\end{table*}

\begin{table*}
  \centering
  \begin{tabular}{@{}l|l|l|c|l|ccc|c@{}}
    \toprule
    Image Encoder & Image Pretraining & Text Pretraining & SS & Pooling & BRCA & NSCLC & RCC & Average \\
    \midrule
    CTP & SSL & In-domain & \multirow{5}{*}{\xmark} & \multirow{5}{*}{topK} & \textbf{0.672} & \textbf{0.700} & \textbf{0.733} & \textbf{0.702} \\
    ViT-S & SSL & In-domain & & & 0.617 & 0.625 & 0.673 & 0.639 \\
    ViT-S & ImageNet & In-domain & & & 0.660 & 0.525 & 0.600 & 0.595 \\
    CTP & None & None & & & 0.535 & 0.520 & 0.297 & 0.451 \\
    ViT-S & None & None & & & 0.500 & 0.510 & 0.290 & 0.433 \\
    \midrule
    CTP & SSL & In-domain & \multirow{5}{*}{\cmark} & \multirow{5}{*}{topK} & 0.615 & \textbf{0.705} & \textbf{0.733} & \textbf{0.684} \\
    ViT-S & SSL & In-domain & & & 0.625 & 0.603 & 0.653 & 0.627 \\
    ViT-S & ImageNet & In-domain & & & \textbf{0.650} & 0.512 & 0.657 & 0.606 \\
    CTP & None & None & & & 0.528 & 0.527 & 0.313 & 0.456 \\
    ViT-S & None & None & & & 0.495 & 0.495 & 0.310 & 0.433 \\
    \midrule
    CTP & SSL & In-domain & \multirow{5}{*}{\xmark} & \multirow{5}{*}{mean} & \textbf{0.620} & \textbf{0.590} & \textbf{0.633} & \textbf{0.614} \\
    ViT-S & SSL & In-domain & & & 0.590 & 0.515 & 0.543 & 0.549 \\
    ViT-S & ImageNet & In-domain & & & 0.615 & 0.510 & 0.580 & 0.568 \\
    CTP & None & None & & & 0.535 & 0.560 & 0.397 & 0.497 \\
    ViT-S & None & None & & & 0.497 & 0.520 & 0.317 & 0.445 \\
    \midrule
    CTP & SSL & In-domain & \multirow{5}{*}{\cmark} & \multirow{5}{*}{mean} & \textbf{0.625} & \textbf{0.590} & \textbf{0.637} & \textbf{0.617} \\
    ViT-S & SSL & In-domain & & & 0.590 & 0.515 & 0.540 & 0.548 \\
    ViT-S & ImageNet & In-domain & & & 0.615 & 0.510 & 0.573 & 0.566 \\
    CTP & None & None & & & 0.542 & 0.562 & 0.393 & 0.499 \\
    ViT-S & None & None & & & 0.495 & 0.520 & 0.320 & 0.445 \\
    \bottomrule
  \end{tabular}
  \caption{\textbf{Pretraining comparison}. Balanced accuracy of different pretraining configurations across all 3 tasks on in-house independent test set. All configurations above use HistPathGPT as the text encoder. \textbf{SS}: spatial smoothing.}
  \label{tab:randsupp}
\end{table*}


\begin{table*}
  \centering
  \begin{tabular}{@{}l|c|c|l|ccc|c@{}}
    \toprule
    Text Encoder \& Pretraining & LIT & SS & Pooling & BRCA & NSCLC & RCC & Average \\
    \midrule
    \multirow{2}{*}{HistPathGPT (In-domain)} & \xmark & \multirow{2}{*}{\xmark} & \multirow{2}{*}{topK} & \textbf{0.672} & 0.700 & 0.733 & 0.702 \\
    & \cmark & & & 0.690 & 0.670 & 0.760 & \textbf{0.707} \\
    \midrule
    \multirow{2}{*}{PubMedBert (Out-of-domain)} & \xmark & \multirow{2}{*}{\xmark} & \multirow{2}{*}{topK} & 0.570 & 0.693 & \textbf{0.777} & 0.680 \\
    & \cmark & & & 0.597 & 0.615 & 0.643 & 0.619 \\
    \midrule
    \multirow{2}{*}{BioClinicalBert (Out-of-domain)} & \xmark & \multirow{2}{*}{\xmark} & \multirow{2}{*}{topK} & 0.660 & \textbf{0.742} & 0.697 & 0.700 \\
    & \cmark & & & 0.575 & 0.623 & 0.547 & 0.581 \\
    \hline
    \midrule
    \multirow{2}{*}{HistPathGPT (In-domain)} & \xmark & \multirow{2}{*}{\cmark} & \multirow{2}{*}{topK} & 0.615 & 0.705 & 0.733 & 0.684 \\
    & \cmark & & & \textbf{0.688} & 0.675 & 0.740 & \textbf{0.701} \\
    \midrule
    \multirow{2}{*}{PubMedBert (Out-of-domain)} & \xmark & \multirow{2}{*}{\cmark} & \multirow{2}{*}{topK} & 0.577 & 0.725 & \textbf{0.760} & 0.688 \\
    & \cmark & & & 0.595 & 0.625 & 0.647 & 0.622 \\
    \midrule
    \multirow{2}{*}{BioClinicalBert (Out-of-domain)} & \xmark & \multirow{2}{*}{\cmark} & \multirow{2}{*}{topK} & 0.660 & \textbf{0.770} & 0.663 & 0.698 \\
    & \cmark & & & 0.600 & 0.635 & 0.543 & 0.593 \\
    \hline
    \midrule
    \multirow{2}{*}{HistPathGPT (In-domain)} & \xmark & \multirow{2}{*}{\xmark} & \multirow{2}{*}{mean} & 0.620 & 0.590 & 0.633 & 0.614 \\
    & \cmark & & & 0.603 & 0.557 & 0.600 & 0.587 \\
    \midrule
    \multirow{2}{*}{PubMedBert (Out-of-domain)} & \xmark & \multirow{2}{*}{\xmark} & \multirow{2}{*}{mean} & 0.585 & 0.650 & \textbf{0.727} & \textbf{0.654} \\
    & \cmark & & & 0.573 & 0.557 & 0.543 & 0.558 \\
    \midrule
    \multirow{2}{*}{BioClinicalBert (Out-of-domain)} & \xmark & \multirow{2}{*}{\xmark} & \multirow{2}{*}{mean} & \textbf{0.672} & \textbf{0.680} & 0.543 & 0.632 \\
    & \cmark & & & 0.607 & 0.575 & 0.533 & 0.572 \\
    \hline
    \midrule
    \multirow{2}{*}{HistPathGPT (In-domain)} & \xmark & \multirow{2}{*}{\cmark} & \multirow{2}{*}{mean} & 0.625 & 0.590 & 0.637 & 0.617 \\
    & \cmark & & & 0.605 & 0.557 & 0.600 & 0.588 \\
    \midrule
    \multirow{2}{*}{PubMedBert (Out-of-domain)} & \xmark & \multirow{2}{*}{\cmark} & \multirow{2}{*}{mean} & 0.587 & 0.650 & \textbf{0.730} & \textbf{0.656} \\
    & \cmark & & & 0.575 & 0.560 & 0.543 & 0.559 \\
    \midrule
    \multirow{2}{*}{BioClinicalBert (Out-of-domain)} & \xmark & \multirow{2}{*}{\cmark} & \multirow{2}{*}{mean} & \textbf{0.675} & \textbf{0.682} & 0.543 & 0.634 \\
    & \cmark & & & 0.613 & 0.577 & 0.533 & 0.574 \\
    \bottomrule
  \end{tabular}
  \caption{\textbf{Locked-image tuning}. Balanced accuracy of models using locked-image tuning across all 3 tasks on in-house independent test set. All configurations above use CTP as the image encoder. \textbf{LIT}: locked-image tuning, \textbf{SS}: spatial smoothing.}
  \label{tab:lit}
\end{table*}

\begin{table*}
  \centering
  \begin{tabular}{@{}p{3cm}|l|c|l||ccc|c}
    \toprule
    Model & Text Encoder \& Pretraining & SS & Pooling & BRCA & NSCLC & RCC & Average \\
    \midrule
    ABMIL (1\% Data) & None &  \multirow{2}{*}{\xmark} & \multirow{2}{*}{attention} & 0.522 & 0.746 & 0.625 & 0.631 \\
    ABMIL (100\% Data) & None & & & 0.863 & 0.907 & 0.946 & 0.905 \\
    \midrule
    \multirow{4}{*}{MI-Zero (Ours)} & HistPathGPT (None) & \multirow{4}{*}{\xmark} & \multirow{4}{*}{topK} & 0.636 & 0.647 & 0.728 & 0.670 \\
    & HistPathGPT (In-domain) & & & 0.642 & 0.717 & 0.832 & \textbf{0.730} \\
    & PubmedBert (Out-of-domain) & & & 0.554 & 0.673 & \textbf{0.871} & 0.700 \\
    & BioclinicalBert (Out-of-domain) & & & \textbf{0.660} & \textbf{0.732} & 0.723 & 0.705 \\
    \midrule
    \multirow{4}{*}{MI-Zero (Ours)} & HistPathGPT (None) & \multirow{4}{*}{\cmark} & \multirow{4}{*}{topK} & 0.639 & 0.675 & 0.736 & 0.683 \\
    & HistPathGPT (In-domain) & & & 0.619 & 0.701 & 0.846 & \textbf{0.722} \\
    & PubmedBert (Out-of-domain) & & & 0.568 & 0.709 & \textbf{0.867} & 0.715 \\
    & BioclinicalBert (Out-of-domain) & & & \textbf{0.677} & \textbf{0.749} & 0.731 & 0.719 \\
    \midrule
    \multirow{4}{*}{MI-Zero (Ours)} & HistPathGPT (None) & \multirow{4}{*}{\xmark} & \multirow{4}{*}{mean} & 0.680 & 0.582 & 0.698 & 0.653 \\
    & HistPathGPT (In-domain) & & & 0.626 & 0.660 & 0.777 & 0.688 \\
    & PubmedBert (Out-of-domain) & & & 0.593 & 0.673 & \textbf{0.806} & 0.691 \\
    & BioclinicalBert (Out-of-domain) & & & \textbf{0.721} & \textbf{0.728} & 0.659 & \textbf{0.703} \\
    \midrule
    \multirow{4}{*}{MI-Zero (Ours)} & HistPathGPT (None) & \multirow{4}{*}{\cmark} & \multirow{4}{*}{mean} & 0.680 & 0.583 & 0.697 & 0.653 \\
    & HistPathGPT (In-domain) & & & 0.625 & 0.660 & 0.776 & 0.687 \\
    & PubmedBert (Out-of-domain) & & & 0.592 & 0.671 & \textbf{0.806} & 0.689 \\
    & BioclinicalBert (Out-of-domain) & & & \textbf{0.722} & \textbf{0.728} & 0.660 & \textbf{0.703} \\
    \bottomrule
  \end{tabular}
  \caption{\textbf{Additional results on TCGA.} Balanced accuracies are shown. We observe similar trends compared to in-house independent test set results. Pretraining the text encoder on unpaired data yields the better results than no pretraining. Spatial smoothing does not yield consistent improvement. TopK pooling performs better than mean pooling for all models.}
  \label{tab:main_tcga}
\end{table*}

\section{Runtime analysis}
\subsection{Comparing runtime of MI-Zero against ABMIL}
We expect MI-Zero inference to outperform SOTA methods that utilize learned attention operators (typically require a forward pass through a multi-layer neural network), after patch embeddings are extracted, MI-Zero only requires a single linear projection to project them into the shared visual-language latent space and computing the cosine similarity scores between each patch and each class prompt (which can be efficiently implemented using matrix multiplication). Note while MI-Zero also needs to build the zero-shot classifier by embedding the class prompts using the text encoder and projecting them into the shared latent space, this step only needs to be computed once for each task and prompt, and can be cached for future use (i.e., the cost is amortized across all samples). 

Using the in-house BRCA subtyping dataset (average bag size is 8767.42 patches per WSI), we benchmarked the run time of topK pooling MI-Zero, which we found to perform consistently well across all tasks. When not considering IO, MI-Zero inference took between 0.70 to 0.75 ms per WSI depending on the K used, which is nearly 2x the speed of ABMIL inference at 1.4 ms per WSI, in line with our expecation. We found on average, building the zero-shot classifier using the text encoder took around 70 ms per prompt (including time to ensemble multiple templates), which represents an additional amortized cost of 0.35 ms per WSI for our test set of 200 WSIs. Lastly, we note that in the current workflows of embedding-based MIL inference, IO (i.e. reading the embeddings from disk to memory and transferring the tensors to the GPU) still represents the primary bottleneck, which took nearly 20ms per WSI on our workstation despite using a fast SSD.

{\small
\bibliographystyle{ieee_fullname}
\bibliography{egbib}
}